%% file: main.tex
\documentclass[10pt,twocolumn,letterpaper]{article}

\usepackage{wacv}
\usepackage{times}
\usepackage{epsfig}
\usepackage{graphicx}
\usepackage{amsmath}
\usepackage{amssymb}

\usepackage{import}
\usepackage{subcaption}
\usepackage{caption}
\usepackage{enumitem}
\usepackage[subtle]{savetrees}
\newcommand{\argmin}{\mathop{\mathrm{argmin~}}}
\newcommand{\argmax}{\mathop{\mathrm{argmax~}}}

\usepackage[pagebackref=true,breaklinks=true,letterpaper=true,colorlinks,bookmarks=false]{hyperref}

\wacvfinalcopy 

\def\wacvPaperID{235} 

\ifwacvfinal\fi
\begin{document}

\title{Improved Embeddings with Easy Positive Triplet Mining}

\author{Hong Xuan, Abby Stylianou and Robert Pless \\
George Washington University\\
{\tt\small \{xuanhong|astylianou|pless\}@gwu.edu}
}

\maketitle
\ifwacvfinal\thispagestyle{empty}\fi

\begin{abstract}
Deep metric learning seeks to define an embedding where semantically similar images are embedded to nearby locations, and semantically dissimilar images are embedded to distant locations. Substantial work has focused on loss functions and strategies to learn these embeddings by pushing images from the same class as close together in the embedding space as possible. In this paper, we propose an alternative, loosened embedding strategy that requires the embedding function only map each training image to the most similar examples from the same class, an approach we call ``Easy Positive'' mining.  We provide a collection of experiments and visualizations that highlight that this Easy Positive mining leads to embeddings that are more flexible and generalize better to new unseen data. This simple mining strategy yields recall performance that exceeds state of the art approaches (including those with complicated loss functions and ensemble methods) on image retrieval datasets including CUB, Stanford Online Products, In-Shop Clothes and Hotels-50K. Code is available at:  \textcolor{magenta}{https://github.com/littleredxh/EasyPositiveHardNegative}
\end{abstract}

\section{Introduction}
Deep metric learning seeks to define an embedding where semantically similar images are embedded to nearby locations, and semantically dissimilar images are embedded to distant locations.  A number of approaches have been proposed for this problem, but many of them learn this embedding by considering triplets of images: an anchor image, a positive image from the same class, and a negative image from a different class.  The network is trained to minimize a loss function that penalizes cases where the anchor-positive image distance is not substantially smaller than the anchor-negative image distance.

When these embedding spaces are learned for the purpose of classification, the evaluation for test data has several steps.  First, images from all classes of interest are mapped into the embedding space.  When a query arrives, it is mapped into the embedding space.  The classification result is the class of whichever image the query was mapped closest to.

\import{figs/}{front_page.tex}

There have been a variety of triplet selection approaches proposed to maximize accuracy on this classification from embedding task. The general triplet selection model (Figure~\ref{fig:frontpage_random}) chooses random anchor-positive pairs, while another approach~\cite{SOP} considers all possible triplets in a batch (Figure~\ref{fig:frontpage_batchall}). These strategies align with the common goal of metric learning, which is to cluster all images from the same class as closely as possible in the embedding space. 

This does not align with the query criteria on some high intra-class variance datasets such as CUB-200 ~\cite{CUB200}, where a query image need not be close to all examples in its class, but rather only close to some example of the class. This suggests that we should optimize the embedding so that images are close to \textit{some} exemplar of their class, but perhaps not \textit{all} exemplars.  Thus we present the idea of ``Easy Positive'' triplet mining (Figure~\ref{fig:frontpage_ephn}), where, for a given anchor image, we find the closest positive example, and optimize to make sure it is closer than negative examples. Figure~\ref{fig:frontpage_birds} motivates this strategy, showing ten images from the `cardinal' class in the CUB-200 dataset~\cite{CUB200}. Many of these images are significantly visually dissimilar, most obviously based on the differences in coloring, but also in background and orientation. Our approach allows these visually dissimilar members of the class to form manifold in the embedding instead of forcing the members to project to the same place in the embedding.

Our specific contributions are:
\begin{itemize}[noitemsep]
    \item Introducing the idea of Easy Positive mining for metric learning and visualizations of the representational flexibility this supports;
    \item Experimental comparisons of Easy Positive with other triplet selection or aggregation approaches; and
    \item Results that demonstrate substantial improvement over other triplet based learning methods, and which improve the state of the art for CARS, CUB, SOP, Fashion, and Hotels-50K compared to all known methods including much more complicated approaches.
\end{itemize}


\section{Background}
There is a large body of work in distance metric learning and its associated loss functions for deep learning, including contrastive loss~\cite{hadsell2006dimensionality}, triplet loss~\cite{deepFeature,facenet,NIPS2005_2795,SOP}, and quadruplet loss~\cite{quadruplet}, as well as more complex variants of triplet loss~\cite{HDC,HTL}, approaches that optimize to project all images from a class to a known location in the embedding space~\cite{Proxy}, and ensemble approaches that combine the output of multiple networks or approaches~\cite{ABE,BIER,DREML}. In this paper we specifically focus on triplet loss and the related N-pair loss~\cite{Npairs}, and different approaches for selecting examples within a batch.

\subsection{Triplet loss}
Triplet loss is trained with triplets of images, $(x_a,x_p,x_n)$, where $x_a$ is an anchor image, $x_p$ is a positive image of the same class as the anchor, and $x_n$ is a negative image of a different class, and the convolutional neural network, $f(\cdot)$, embeds the images on a unit sphere, $(f(x_a), f(x_p), f(x_n))$.  The target is to learn an embedding such that the anchor-positive pair are at least closer together than the anchor-negative pair by some margin: $d_{ap}-d_{an}>margin$, where $d_{ap}=\left \| f(x_a)-f(x_p) \right \|_2$. The loss is then defined as:
\[L=max(0, d_{ap}-d_{an}+margin)\] 

\subsection{N-pair and NCA Loss}
N-pair loss~\cite{Npairs} adds more negative examples into triplets and turns the triplet into an N-tuplet,  $(x_a,x_p,x_{n_1},...,x_{n_i})$. The convolution neural network embeds the images on a unit sphere, $(f(x_a),f(x_p),f(x_{n_1}),...,f(x_{n_i}))$. The authors additionally modify the standard triplet loss function from a margin based loss to an NCA loss~\cite{nca}, avoiding the selection of the margin hyper-parameter and more efficiently pushing positives and negative to be far apart. The loss is defined as:
\[L=-log\frac{\exp{(f(x_a)^{\intercal}f(x_p))}}{\exp{(f(x_a)^{\intercal}f(x_p))}+\sum_{i}exp{(f(x_a)^{\intercal}f(x_{n_i}))}}\] 

\subsection{Triplet Mining}
In order to construct a triplet for a particular anchor image $x_a$, we must select a positive image, $x_p$, from the same class and a negative image, $x_n$, from a different class. In a dataset set with $N$ training images, there are $O(N^3)$ possible triplets, many of which do not help the training converge (e.g., triplets where  $d_{an} >> d_{ap}$). It is important for fast convergence of the triplet based networks to then construct only the most useful triplets. This triplet construction can either occur \textit{offline}, selecting triplets from the entire training dataset after each training snapshot, or \textit{online}, selecting triplets from within each batch in the training. The details of triplet mining will be discussed in further detail in Section~\ref{sec:selection}.

\subsection{Drawbacks of Existing Approaches}
One of the key problems with many existing metric learning approaches is that, when trained only with class labels for supervision, they learn an embedding that pushes together all examples of a class regardless of their semantic similarity. A natural way to embed those examples is to project them on several clusters or a manifold instead of a point to tolerant the semantic differences within a class, as seen in Figure~\ref{fig:frontpage_birds}, or mislabeled data. 

The most significant work to address this issue of over-clustering in deep metric learning introduced the concept of `magnet loss'~\cite{magnet}, where classes are split into clusters using k-means, and points within a cluster are pushed together, but clusters within a class are allowed to spread out. This approach, however, requires the computation and book-keeping of the clusters, and requires regular pauses during training to offline re-compute the clusters within a class using the current representation.

In this paper, we show that our simple, online triplet selection approach focusing on only the most similar positive example is tolerant to high intra-class variance, forming manifold embedding, avoiding over-clustering of the embedding space, and generalizes well to unseen classes.

\import{figs/}{usage1.tex}

\section{Strategies for Triplet Selection}
\label{sec:selection}
In our training, each batch of $N$ images contains $n$ examples from $c$ classes, randomly selected from the training data. Throughout the paper, we refer to the $n$ examples per class as a group. We review different possible online mining strategies to select the most useful examples within a batch both from the same class and from different classes to an anchor image.

\subsection{Hard Negative Mining}
Hard negative examples are the most similar images which have a different label from the anchor image.
\[ x_{hn} = \argmin_{x:C(x)\neq C(x_a)} d(f(x_a),f(x)) \]

Many works have discussed the benefit of hard negative mining in constructing triplets that produce useful gradients and therefore help triplet loss networks converge quickly~\cite{HermansBeyer2017Arxiv,facenet,discriminativeLearning}. In ~\cite{facenet}, the authors additionally propose the concept of Semi-Hard Negative mining, which chooses a anchor-negative pair that is farther than the anchor-positive pair, but within the margin, and so still contributes a positive loss. They demonstrate that using these Semi-Hard Negatives achieves superior performance to networks trained with random or hard negatives. In this paper, given the feature for an anchor image $x_a$ and its positive example $x_p$, we define their Semi-Hard Negative: 
\[ x_{shn} = \argmin_{x:\substack{C(x)\neq C(x_a) \\ d(f(x_a),f(x))>d(f(x_a),f(x_p))}}  d(f(x_a),f(x)) \]

\subsection{Easy Negative Mining}
Easy negative examples are the least similar images which have the different label from the anchor image.
\[ x_{en} = \argmax_{x:C(x)\neq C(x_a)} d(f(x_a),f(x)) \]

This condition is not useful in triplet construction, as it will not produce useful gradients for updating the model.

\subsection{Hard Positive Mining}
Hard positive examples are the least similar images which have the same label to as anchor image.
\[ x_{hp} = \argmax_{x:C(x) = C(x_a)} d(f(x_a),f(x)) \]

In~\cite{HermansBeyer2017Arxiv}, the authors show that hard positive examples increase clustering within a class. The authors also empirically demonstrate that hard positive mining is not universally suitable for all datasets. In Section~\ref{sec:results}, we show that the primary problem with hard positive mining is actually related to the number of images per class in the batch (if there is a large number of examples per class in a batch, the likelihood that the hardest anchor-positive pair in a batch are very dissimilar increases).

\import{figs/}{usage2.tex}

\subsection{Easy Positive Mining}

Our solution to address the over-clustering of the embedding space and to keep intra-class variance in real data is to compute the loss using the easiest positive pair per class in the batch. The easy positive examples are the most similar images that have the same label as the anchor image:
\[ x_{ep} = \argmin_{x:C(x) = C(x_a)} d(f(x_a),f(x)) \]

This selection will more likely to push two close positives together and less likely to push two far away positives together. Therefore, it can maintain the intra-class variance and allow classes to have manifold structure in the embedding space and and help reduce the over-clustering problem when an embedding must map dramatically different images to the same place.  Figure~\ref{fig:usage_train} shows that training data is less clustered using the easy-positive mining than when using existing approaches and Figure~\ref{fig:tsne} shows that training examples are embedded on flexible manifold instead of points.  

Additionally, this approach seems to generalize better to unseen data than the other approaches. Figure~\ref{fig:usage_test} shows where testing data embeds with respect to the closest points in the training data.  Existing approaches tend to embed test data from new classes close to where training data was embedded, which Easy Positive approaches spread the new data out more.

\section{Easy Positive Triplet Loss}
Given the Easy Positive mining described above, we can derive our loss function when selecting easy positive examples and different strategies for selecting negative examples. In our approach we follow the standard practice of mapping the output of our convolutional neural network onto a unit sphere.  Then we can compute the similarity of a feature vector for an anchor image $f(x_a)$ and its closest positive $f(x_{ep})$ as the dot product of these vectors: $f(x_a)^\top f(x_{ep})$

Given an anchor image and its feature vector $f(x_a)$, we find both the easy positive $f(x_{EP})$ in the batch and all possible negative examples, and define the Easy Positive (EP) loss as:

\[L_{EP} = -log\frac{\exp{(f(x_a)^{\intercal}f(x_{ep}))}}{\exp{(f(x_a)^{\intercal}f(x_{ep}))}+\sum_{i}exp{(f(x_a)^{\intercal}f(x_{n_i}))}}\] 

We can also find the Easy Positive $f(x_{ep})$, Hard Negative $f(x_{hn})$ and Semi-Hard Negative $f(x_{shn})$ examples for $f(x_a)$ and define the Easy Positive Hard Negative (EPHN) loss and  Easy Positive Semi-Hard Negative (EPSHN) loss as:

\[ L_{EPHN} = -log\frac{\exp{(f(x_{a})^{\intercal}f(x_{ep}))}}{\exp{(f(x_{a})^{^\intercal}f(x_{ep}))}+\exp{(f(x_{a})^{^\intercal}f(x_{hn}))}} \]

\[ L_{EPSHN} = -log\frac{\exp{(f(x_{a})^{\intercal}f(x_{ep}))}}{\exp{(f(x_{a})^{^\intercal}f(x_{ep}))}+\exp{(f(x_{a})^{^\intercal}f(x_{shn}))}} \]

We can additionally compare our Easy Positive triplet losses with the hard positive mining approaches. In these cases, the loss functions, defined as Hard Positive (HP) and Hard Positive Hard Negative (HPHN) are computed as follows:

\[L_{HP} = -log\frac{\exp{(f(x_a)^{\intercal}f(x_{hp}))}}{\exp{(f(x_a)^{\intercal}f(x_{hp}))}+\sum_{i}exp{(f(x_a)^{\intercal}f(x_{n_i}))}}\] 

\[ L_{HPHN} = -log\frac{\exp{(f(x_{a})^{\intercal}f(x_{hp}))}}{\exp{(f(x_{a})^{^\intercal}f(x_{hp}))}+\exp{(f(x_{a})^{^\intercal}f(x_{hn}))}} \]

When the group size is 2, Easy Positive and Hard Positive become random Positive, making $L_{EP}$ and $L_{HP}$ equivalent to N-pair loss.

\section{Experiment}
We calculate Recall@K to measure retrieval quality. To compute Recall@K, we first embed all query set and gallery set images to the unit hyper-sphere and calculate pair-wise cosine similarity between these two sets. For each query image, we retrieve the images with the K highest similarity scores from the gallery set. A recall score is 1 if at least one image of the K retrieved images have the same label as the query image, and 0 if none of the K retrieved images have the same label as the query image. Recall@K is the average of the recall score for all queries.
\import{figs/}{mining.tex}
\subsection{Implementation}
All tests are run on the PyTorch platform~\cite{pytorch}, using the GoogleNet~\cite{googlenet} and ResNet18 and ResNet50~\cite{resnet} architectures, pre-trained on ILSVRC 2012-CLS data~\cite{ILSVRC15}. Training images are re-sized to 256 by 256 pixels.  We adopt a standard data augmentation scheme (random horizontal flip and random crops padded by 10 pixels on each side). For pre-processing, we normalize the images using the channel means and standard deviations. All networks are trained using stochastic gradient descent (SGD) with 40 epoches. We set initial learning rate 0.0005 for CAR, SOP and In-shop cloth dataset and 0.0001 for CUB dataset, and divided by 10 after 20 and 30 epochs. The loss function is based on NCA which has a single parameter, temperature~\cite{heatedup}, and in all cases we set this parameter to 0.1.

On all datasets we train using a batch size of 128.  Batches are constructed with a fixed number $n$ examples per class by adding classes until the batch is full. When a class has fewer than $n$ examples, we use all the examples from the class.  If this leads to a case where the last class in the batch does not have space for $n$ images, we just include enough images to fill the batch.

\subsection{Datasets}
The {\bf CUB200} dataset~\cite{CUB200} contains 200 classes of birds with 11,788 images. We split the first 100 classes for training (5,864 images) and the rest of the classes for testing (5,924 images). In the training set, the maximum, minimum, mean and standard deviation of the number of images in each class is 60, 41, 58.6 and 3.5.

The {\bf CAR196} dataset~\cite{CAR196} contains 196 classes of cars with 16,185 images. We use the standard split with the first 98 classes for training (8,054 images) and the rest of the classes for testing (8,131 images). In the training set, the maximum, minimum, mean and standard deviation of the number of images in each class is 97, 59, 82.2 and 7.2.

The {\bf In-Shop Clothes Retrieval (In-Shop)} dataset~\cite{ICR} contains 11,735 classes of clothing items with 54,642 images. Following the settings in~\cite{ICR}, only 7,982 classes of clothing items with 52,712 images are used for training and testing. 3,997 classes are for training (25,882 images) and 3,985 classes are for testing (28,760 images). The test set are partitioned to query set and gallery set, where query set contains 14,218 images of 3,985 classes and gallery set contains 12,612 images of 3,985 classes. In the training set, the maximum, minimum, mean and standard deviation of the number of images in each class is 162, 1, 6.5 and 6.4.

The {\bf Stanford online products (SOP)} dataset~\cite{SOP} contains 22,634 classes with 120,053 product images. We use 11,318 classes for training (59,551 images) and other 11,316 classes are for testing (60,502 images). In the training set, the maximum, minimum, mean and standard deviation of the number of images in each class is 12, 2, 5.3 and 3.0.

The {\bf Hotel-50K (Hotel)} training dataset~\cite{hotels50k} contains 50,000 hotel classes with 1,027,871 images of hotel rooms within each hotel. In the training set, the maximum, minimum, mean and standard deviation of image size in each class is 266, 2, 20.5, 13.6. The testing dataset consists of 17,954 images from 5,000 hotels represented in the training set.

In the CUB, CAR and SOP datasets, both the query set and gallery set refer to the testing set. During the query process, the top-K retrieved images exclude the query image itself. In the In-Shop dataset, the query set and gallery set are predefined by the original paper. In the Hotel dataset, the training set is used as the gallery for all query images in the test set.

\import{figs/}{Distr.tex}
\import{figs/}{tsne.tex}
\import{table/}{CUBCAR_model.tex}
\import{table/}{SOTA.tex}
\import{table/}{Hotels.tex}

\section{Results}
\label{sec:results}
We show three classes of experiments.  The first gives an extensive comparison between different mining approaches on the CAR dataset, comparing different parameter settings and showing visualizations that offer intuitions about how different mining strategies affect the embedding. The second set of experiments shows comparisons between different algorithm performance for different network architectures.  The third experiment compares our algorithm to a wider variety of algorithms on a larger collection of datasets.

\subsection{Comparative Study Using the CAR Dataset}
In the first set of experiments we explore different mining strategies using the CAR dataset~\cite{CAR196}, and specific explore the effect of the number of images per class in each batch.  The CAR dataset has a large number of examples per class making it possible to test with large numbers of images per class.

\paragraph{Impact of Group Size.} We train an embedding network on the CAR dataset, using a ResNet-18 architecture for 6 different mining strategies.  In all cases, we train with a batch size of 128. Figure~\ref{fig:mining} shows the impact of the group size, $n$ (the number of examples per class that is included per batch) on the Recall@1 performance.

When $n=2$, Easy Positive (EP), Hard Positive (HP) and Batch All (BA) have only 2 representatives per class, and each of these approaches use all negatives in the batch, so these three approaches are all the same in this case (and are the equivalent to the standard N-pair approach).  Similarly, Hard Positive Hard Negative (HPHN) is the same as Easy Positive Hard Negative (EPHN) when there are only two examples per class. 

When $n$ increases, the number of examples per class grows, so the easiest positive image is likely to become more similar and the hardest positive is likely to become less similar.  Our results show an increase in performance for all the Easy Positive approaches and a decrease for the hard positive approaches.  When $n>16$ the performance drops for most methods; we believe this is because for large $n$ there are relatively few different classes per batch, and when all negatives examples are drawn from fewer classes they have less variation.

The Hard Positive and Batch All approaches explicitly focus on creating constraints to pull together all examples in a group, which has the impact of forcing all the points in the class to be close.
In contrast, EP, EPHN and EPSHN only constrain the most similar examples per group, and do not force all elements of the class into a cluster.

For the remainder of this section we drop results about about Hard  Positive Hard Negative and just use the results from the group size of 16. We also include the N-pair approach, which is equivalent to the best version of Batch All and Hard Positive (group size=2).

\paragraph{Usage of the Embedding Space.}Figure~\ref{fig:usage_train} illustrates the impact of Easy Positive mining by showing the distribution of similarities between points from the same class.  The points from the same class are much more spread out when the embedding is based on the Easy Positive mining, while the other approaches create tight clusters where all points from the same class are very similar. 

For these six algorithms, Figure~\ref{fig:simdistr} gives another visualization of how the points are distributed, showing for each data point the similarity to the most similar image from the same class and the most similar image from a different class.  The structure of this plot is that if a point is below the $y=x$ line, then to most similar image is from the same class, and would be classified correctly if it was a query image, and we have color coded those points blue.

The left side of this plot includes Batch All, N-pair and Hard Positive mining.  These algorithms give tight clusters and this is visible in these plots for training data (the left-most column of the plot), where all points are very similar to images from the same class, or far to the right on the x-axis.  In contrast, the Easy Positive approaches (the right-side of the figure) are more spread out.  

On testing data, Batch All, N-pair and Hard Positive mining all have significantly larger distances to the closest example than in the training data (there points are less concentrated on the far right along the x-axis).

In contrast, while the Easy Positive approaches have lower similarities to the closest training points, the embedding generalizes better for new classes, giving distributions on test data of distances to nearby positive and negative examples that are similar to the training distributions; we believe this is why we get better results.

\paragraph{Visualization of Embedding Space.} Finally, Figure~\ref{fig:tsne} shows a t-SNE embedding comparing N-pair with Easy Positive Semi-Hard Negative.  Embedding the training points with N-pair leads to tight clusters, while the Easy Positive Semi-Hard Negative embeds classes to be more spread out and even sometimes disjoint (addressing the fact that sometimes classes have multiple modes, like the birds in Figure~\ref{fig:frontpage_birds}).

The embedding of test data points is also interesting.  On the right side of Figure~\ref{fig:tsne}, we show the testing and training embedding at the same time; the red points are the embedded training data (that were shown on the left), and the blue points are the embedded testing data.  On the EPSHN data the testing points are more spread out, and clusters are rarely mapped on top of the training clusters; while the N-pair training points are less spread out and often mapped exactly on top of training categories.  While this does not necessarily impact testing accuracy, it indicates that the embedding has not learned a representation that generalizes to new classes.

\subsection{Comparison Across Architectures}
In the second set of experiments, we train embedding networks on the CUB and CAR dataset and compare our Easy Positive Semi-Hard Negative approach with triplet, N-pair and Proxy loss, using an output embedding dimension size of 64. We compare the results of these approaches across the GoogleNet, Resnet18 and Resnet50 network architectures in Table~\ref{table:CUBCAR}, demonstrating the superiority of our approach to these other comparably simple embedding approaches across all networks for CUB, and for both Resnet architectures for CAR.

\subsection{Comparison with State of the Art}
In the third set of experiments, we compare the the best reported results for several more complex state of the art embedding approaches, including more complex triplet loss approaches~\cite{HDC,HTL} and ensemble based approaches~\cite{BIER,ABE,DREML}. Our embeddings are trained with ResNet50 and an output embedding size of 512. For CUB and CAR, the optimal group size is 16, while the SOP, In-shop and hotel datasets have fewer examples per class, and therefore perform best with a group size of 4.

In Table~\ref{table:SOTA}, the Easy Positive Semi-Hard Negative approach achieves a new record on the CUB dataset, which contains high intra-class variance as seen in Figure~\ref{fig:frontpage_birds}, improving over even significantly more complex ensemble methods such as ABE~\cite{ABE} and DREML~\cite{DREML}. On the CAR dataset, our result is comparable to the ensemble methods. We additionally achieve state of the art results for In-shop and Hotels-50K, and achieve the best reported Recall@1 for the SOP dataset. In the original Hotels-50K~\cite{hotels50k} paper, the authors specifically cite high intra-class variance as a challenge of their dataset. Our EPSHN method doubles the accuracy of their original approach, which uses the Batch All triplet selection strategy and is trained with ResNet50.

\section{Discussion}
The standard definition of distance metric learning is to create a function so that all images from class are mapped to similar locations and images from different classes are judged to be different.

This criteria does not align well with data from natural classes; for example, Figure~\ref{fig:frontpage_birds} shows the Cardinal category in the cub dataset that has at least two semantic clusters (colorful male birds and brown females).  Even within those semantic classes, there may be value in explicitly matching to birds on a branch or birds on the ground.  We posit that there is value in a distance metric learning approach that matches images to the most semantically similar examples, without needing all images to be similar. At query-time, recall accuracy typically depends only on the label of the most similar image, so a metric learning approach that optimizes for this condition fits better than a metric learning approach that requires all images in a class to be similar.

In this paper we show that the simple change of concentrating on easy positive examples within a batch improves performance across a wide range of datasets and outperforms all published results on large datasets (Stanford Online Products, In-shop Clothes, and Hotels-50K), including quite recent and interesting ensemble based methods.

{\small
\bibliographystyle{ieee}
\bibliography{main}
}

\end{document}

%% file: figs/front_page.tex
\begin{figure}[t]
    \centering
    \begin{subfigure}[b]{.32\columnwidth}
    \captionsetup{justification=centering}
    \begin{center}
	    \includegraphics[width=0.99\columnwidth]{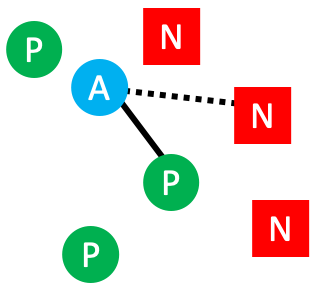}
	    \caption{\label{fig:frontpage_random}Random 
	    \vspace{\baselineskip}
	    }
	 \end{center}
	 \end{subfigure}
    \begin{subfigure}[b]{.32\columnwidth}
    \begin{center}
	    \includegraphics[width=0.99\columnwidth]{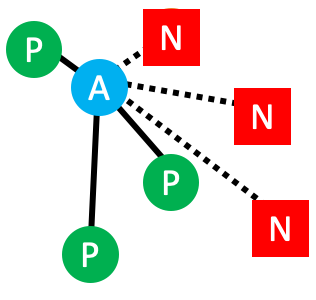}
	    \caption{\label{fig:frontpage_batchall}All Positives\\All Negatives}
	 \end{center}
	 \end{subfigure}
    \begin{subfigure}[b]{.32\columnwidth}
    \begin{center}
	    \includegraphics[width=0.99\columnwidth]{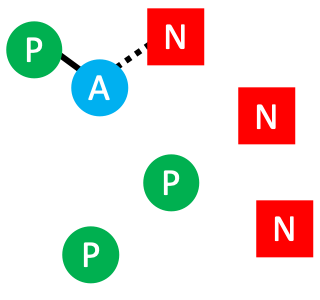}
	    \caption{\label{fig:frontpage_ephn}Easiest Positive\\Hardest Negative}
	 \end{center}
	 \end{subfigure}
	 \begin{subfigure}[]{\columnwidth}
	    \vspace{.25cm}
	    \centering
	    \includegraphics[width=.19\textwidth]{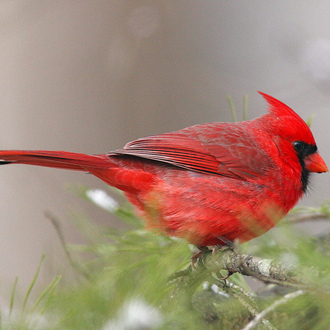}\includegraphics[width=.19\textwidth]{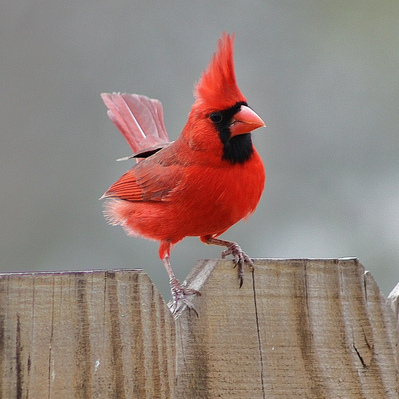}\includegraphics[width=.19\textwidth]{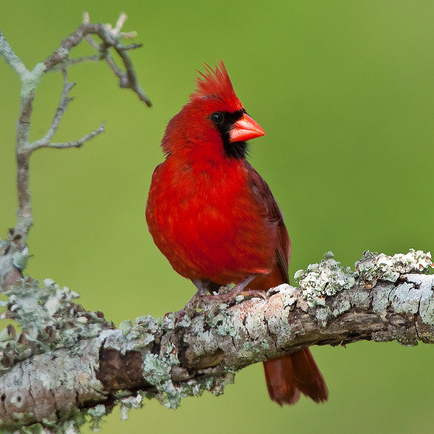}\includegraphics[width=.19\textwidth]{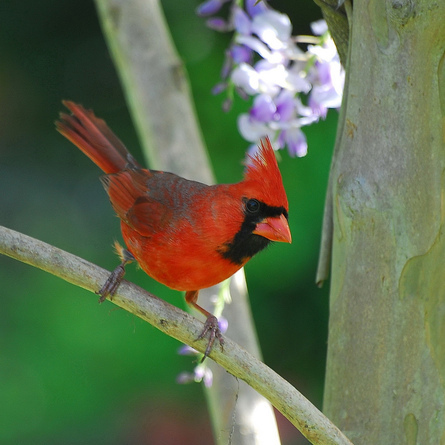}\includegraphics[width=.19\textwidth]{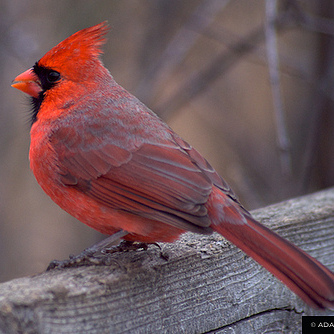}\\
	    \includegraphics[width=.19\textwidth]{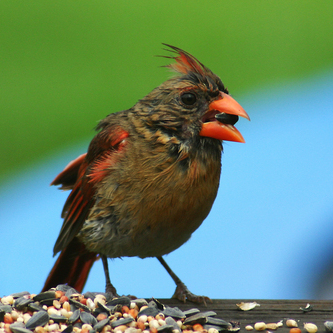}\includegraphics[width=.19\textwidth]{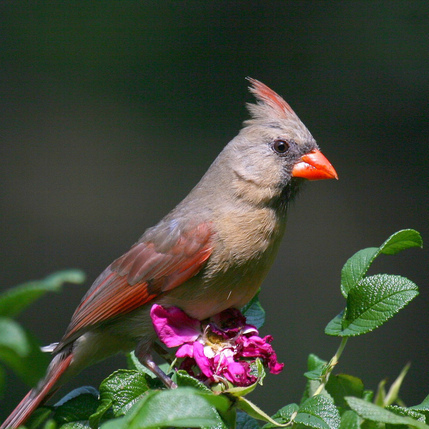}\includegraphics[width=.19\textwidth]{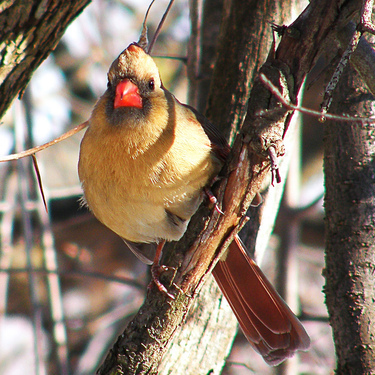}\includegraphics[width=.19\textwidth]{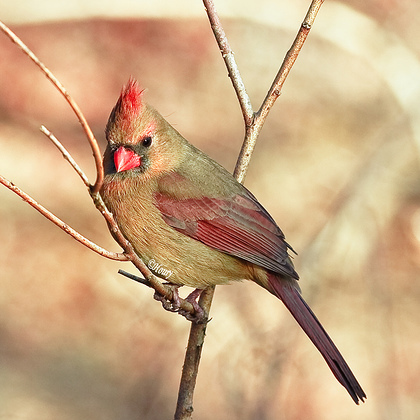}\includegraphics[width=.19\textwidth]{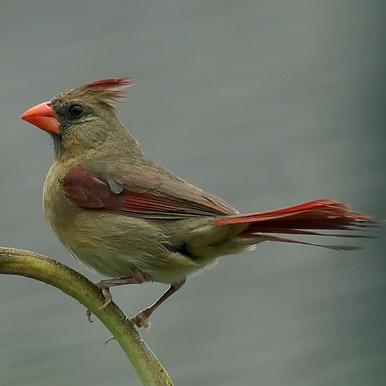}
	    \caption{\label{fig:frontpage_birds}Ten images of the `cardinal' class (id=17) from CUB-200.}
	 \end{subfigure}
	 \caption{Illustration of different triplet selection strategies for an anchor image (blue circle) in a batch. Generic triplet loss (a) randomly selects a positive example (green circle, solid line) and a negative example (red square, dashed line). Batch All~\cite{SOP} (b) considers all possible positive and negative examples. Our approach (c) considers the most similar positive and most similar negative example. The focus on the most similar positive example aligns with the embedding query criteria, where a query example need not be close to all possible examples from the class, but only the most visually similar example. This is motivated by the images in (d), which are all from the `cardinal' class in the CUB~\cite{CUB200} dataset, but many of which have significantly different visual appearances.}
	 \label{fig:front_page}
\end{figure}

%% file: figs/usage1.tex
\begin{figure}[t]
    \centering
    \includegraphics[width=0.99\columnwidth]{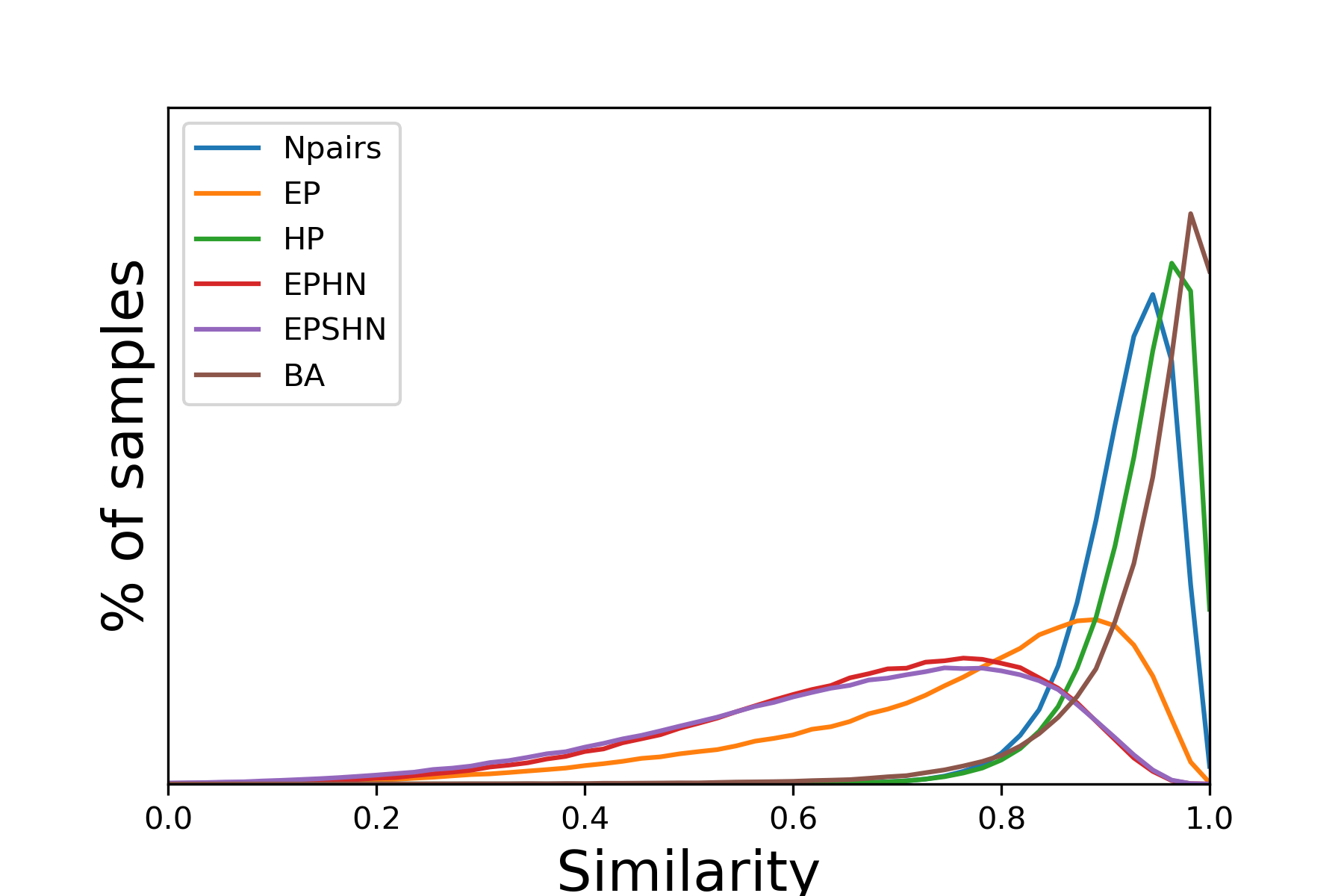}
    \caption{The distribution of similarities between \textbf{all pairs} from the same class in the training data. Batch All (BA), N-Pair, and Hard Positive (HP) all tightly cluster the training data, while the Easy Positive approaches cluster much less tightly.}
	\label{fig:usage_train}
\end{figure}


%% file: figs/usage2.tex
\begin{figure}[t]
    \centering
    \centering
	\includegraphics[width=0.99\columnwidth]{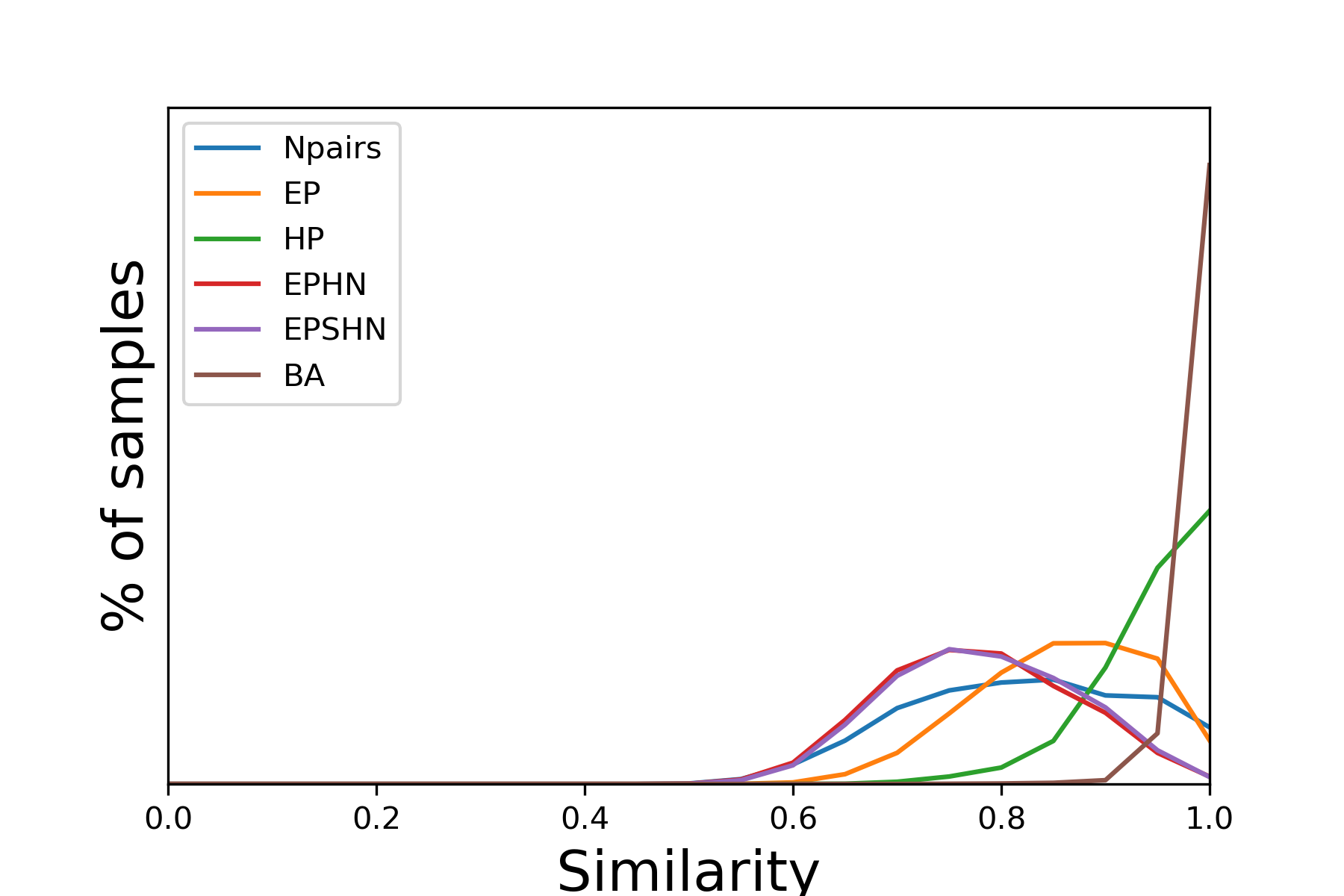}
	\caption{The distribution of similarities between points in the test dataset and their \textbf{closest} point in the training dataset. Batch All (BA), N-Pair, and Hard Positive (HP) all learn an embedding where testing points from new, unseen classes are mapped closely to training data, indicating that they have not learned a representation that differentiates well between training data and unseen testing data. By comparison, all of the Easy Positive approaches map test data much less closely particular training examples.}
	\label{fig:usage_test}
\end{figure}

%% file: figs/mining.tex
\begin{figure}[t]
    \centering
	\includegraphics[width=0.99\columnwidth]{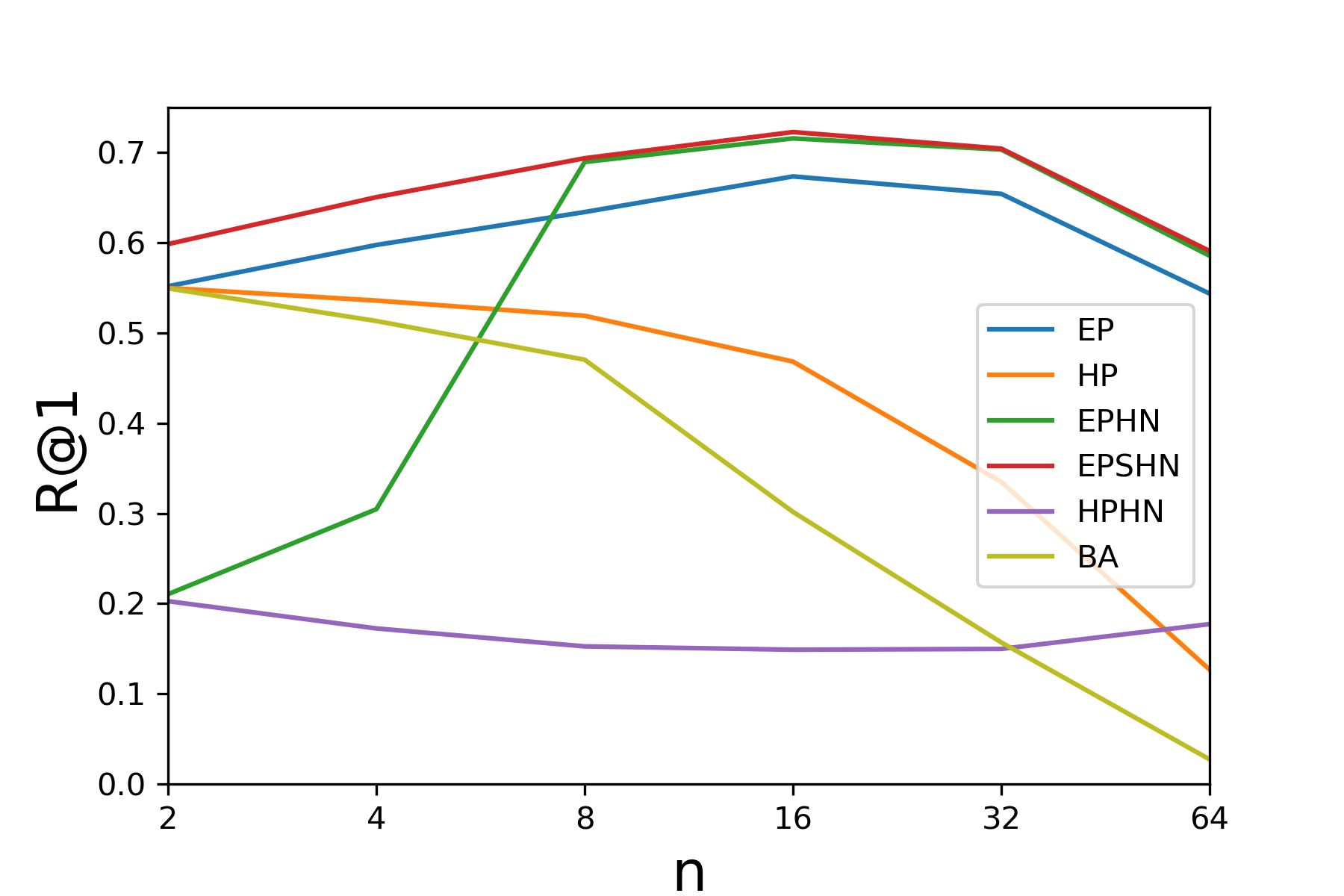}
	\caption{Comparison of Recall@1 for different triplet mining strategies as a function of group-size, $n$ (the number of images from the same class in a batch).}
	\label{fig:mining}
\end{figure}

%% file: figs/Distr.tex
\begin{figure*}[t]
    \begin{subfigure}[b]{.49\textwidth}
    \centering
	\includegraphics[width=0.49\textwidth]{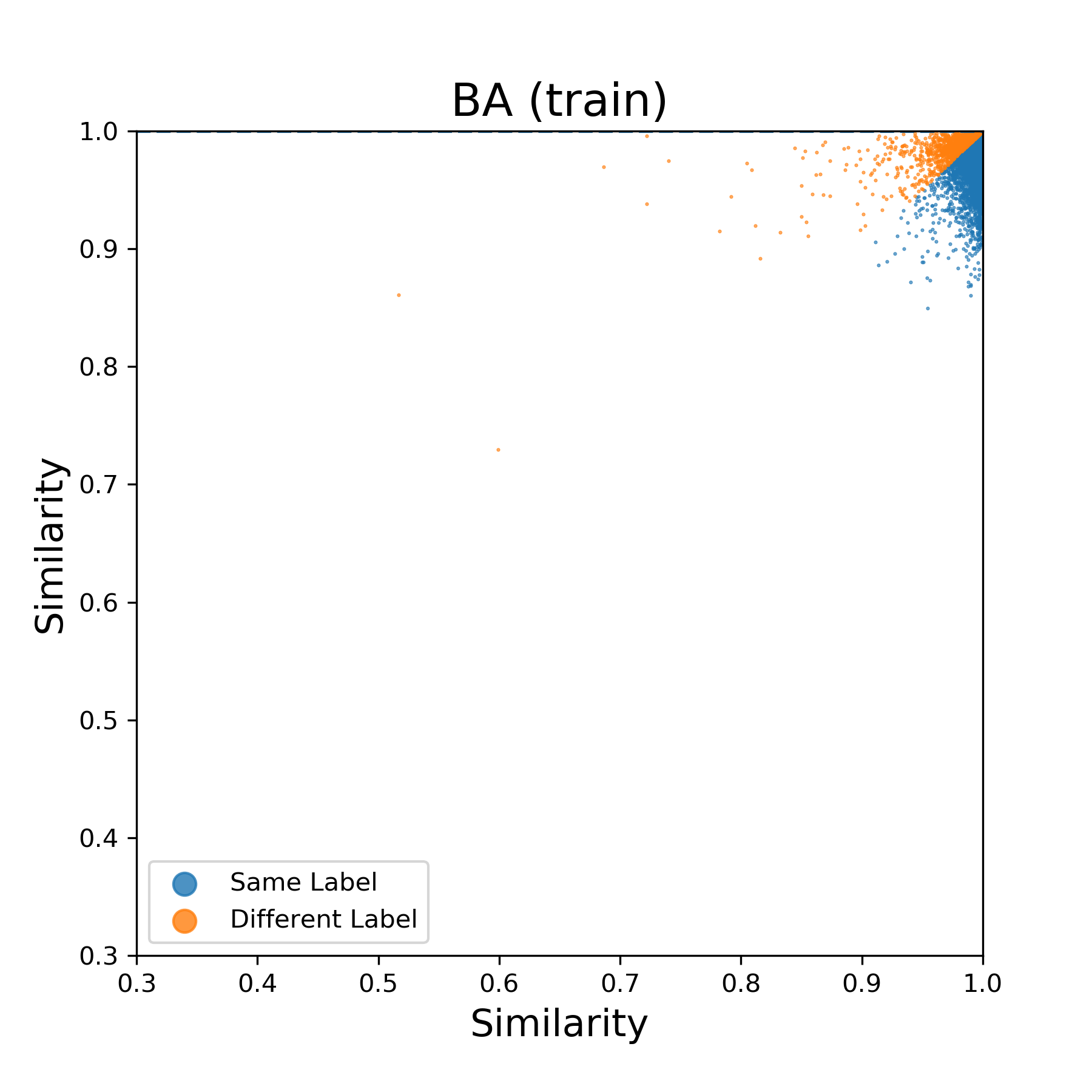}
	\includegraphics[width=0.49\textwidth]{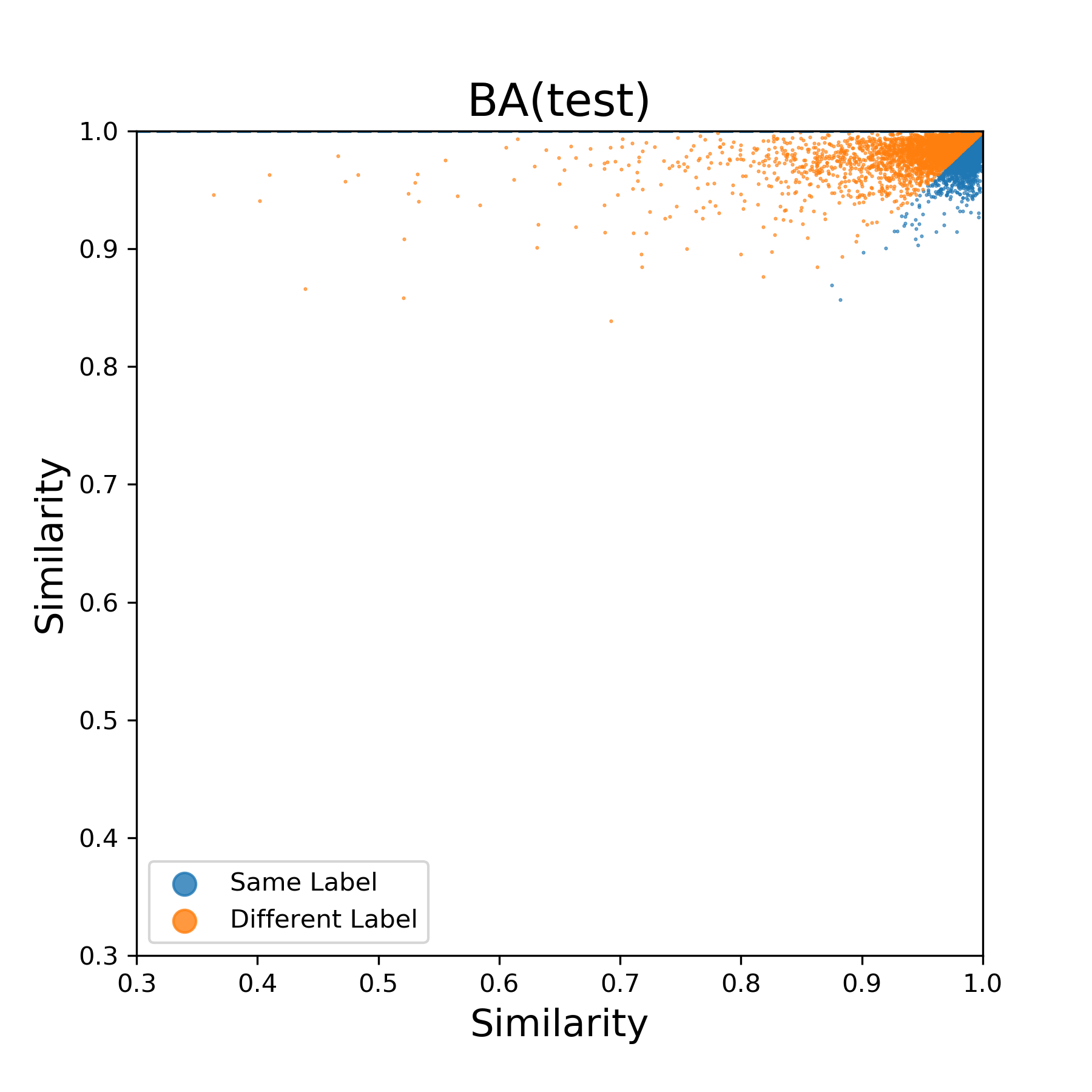}
	\includegraphics[width=0.49\textwidth]{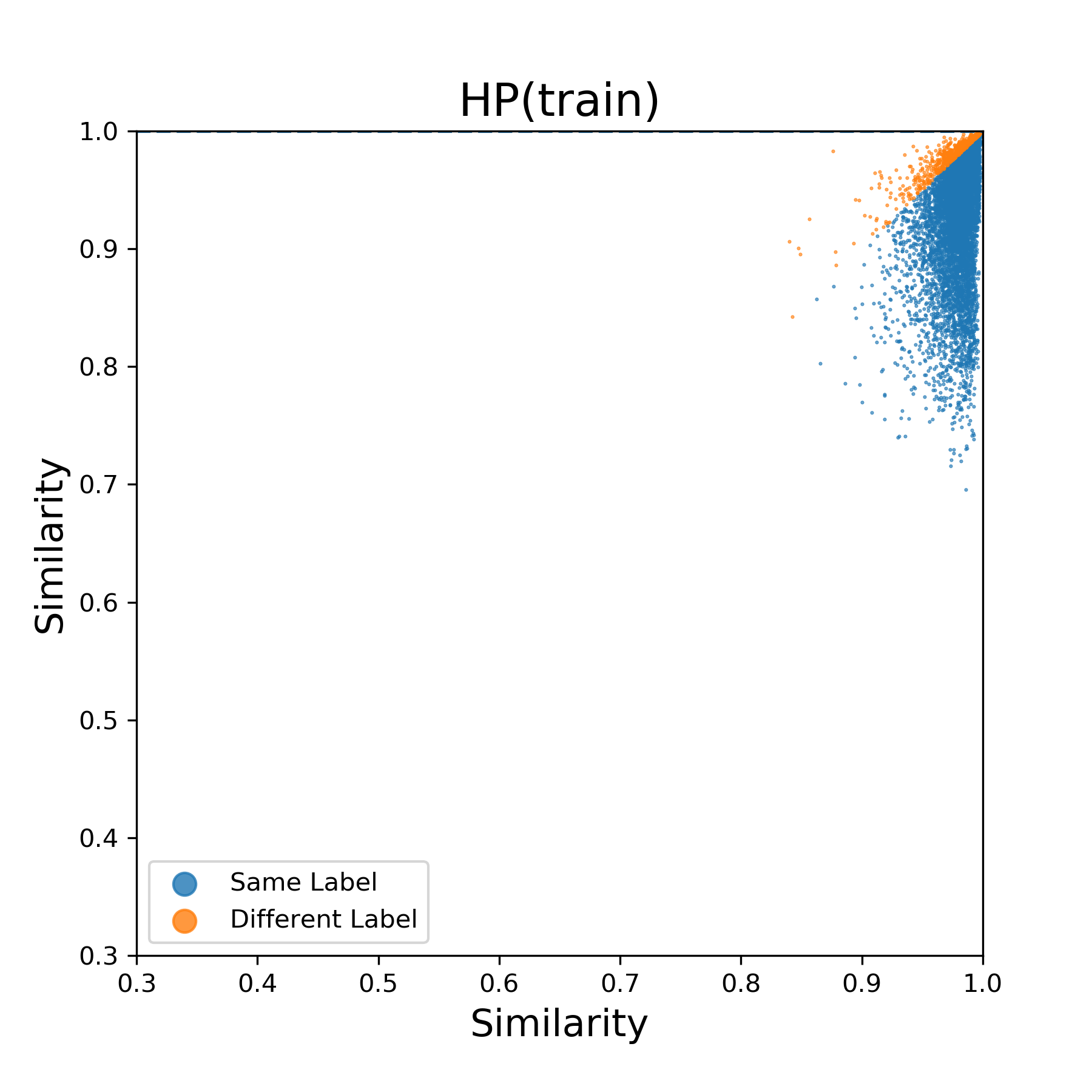}
	\includegraphics[width=0.49\textwidth]{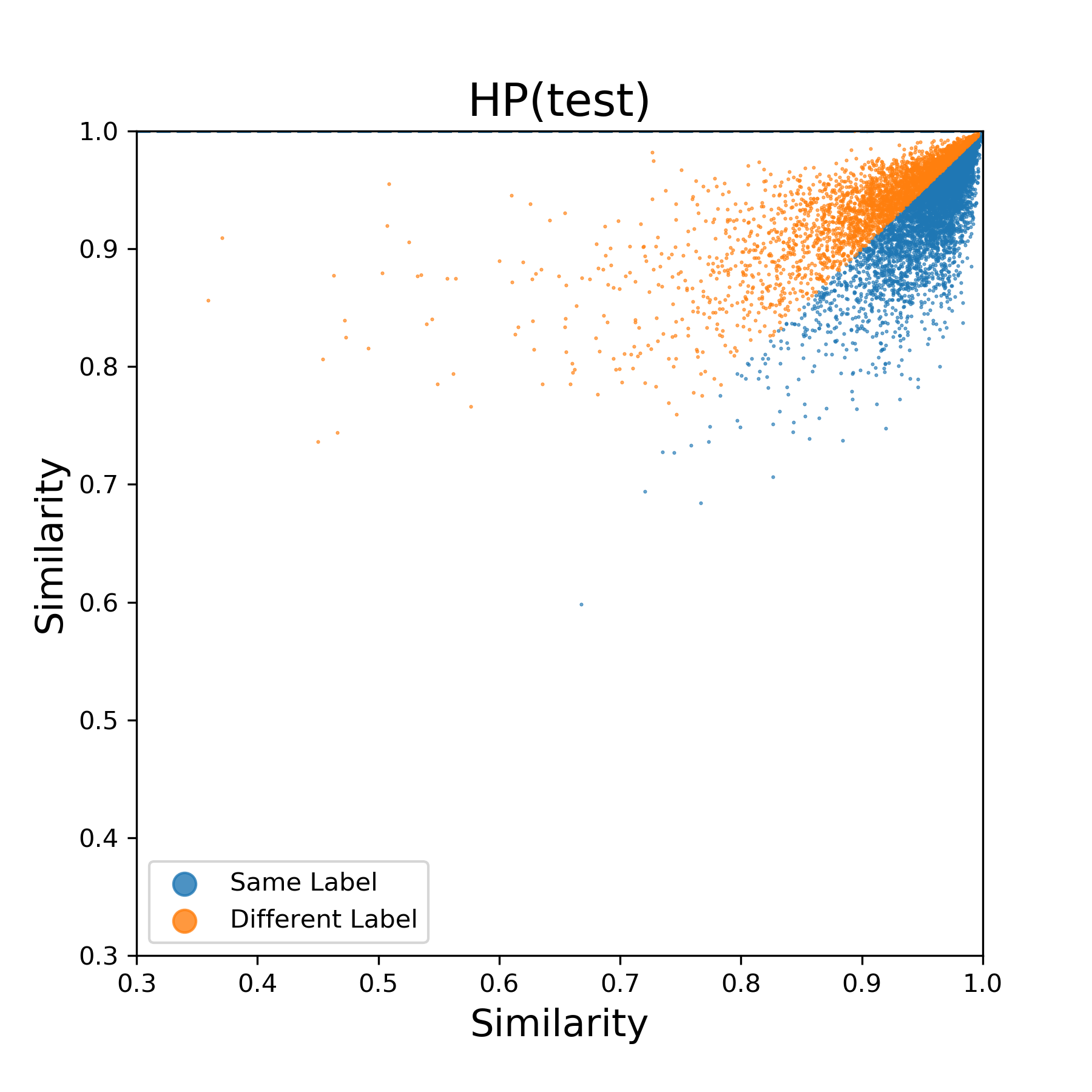}
	\includegraphics[width=0.49\columnwidth]{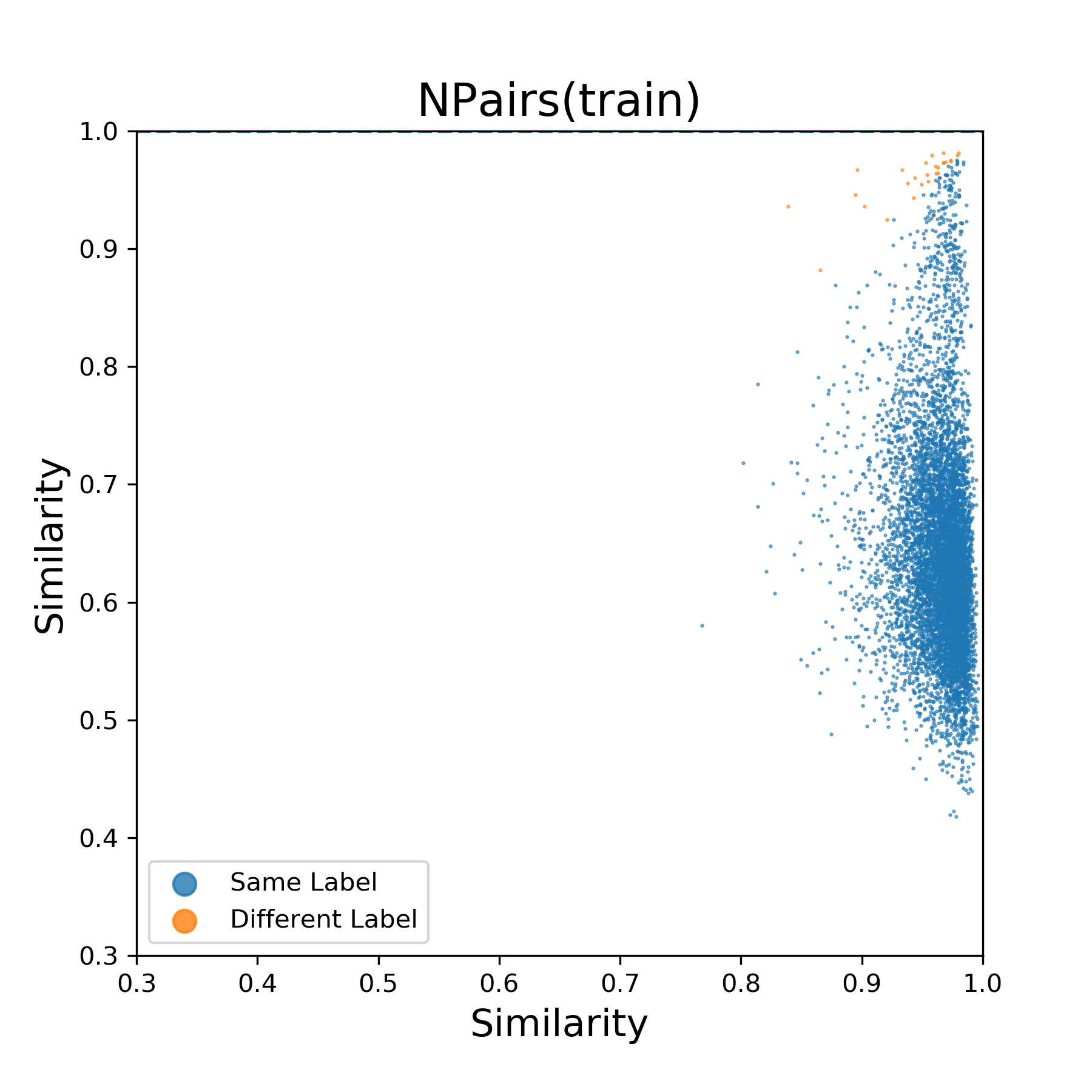}
	\includegraphics[width=0.49\columnwidth]{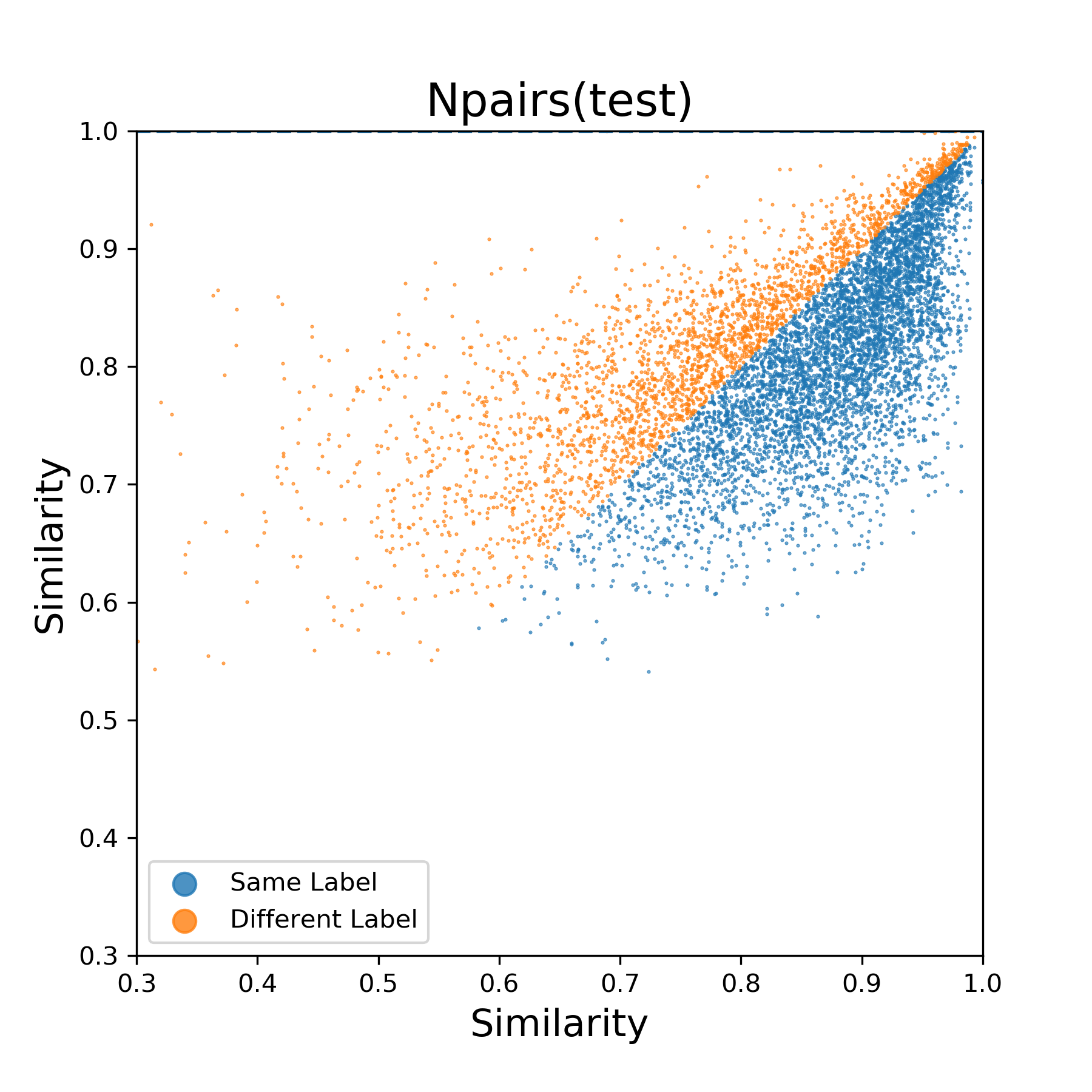}
	\end{subfigure}
	\hfill\vline\hfill
	\begin{subfigure}[b]{.49\textwidth}
	\includegraphics[width=0.49\textwidth]{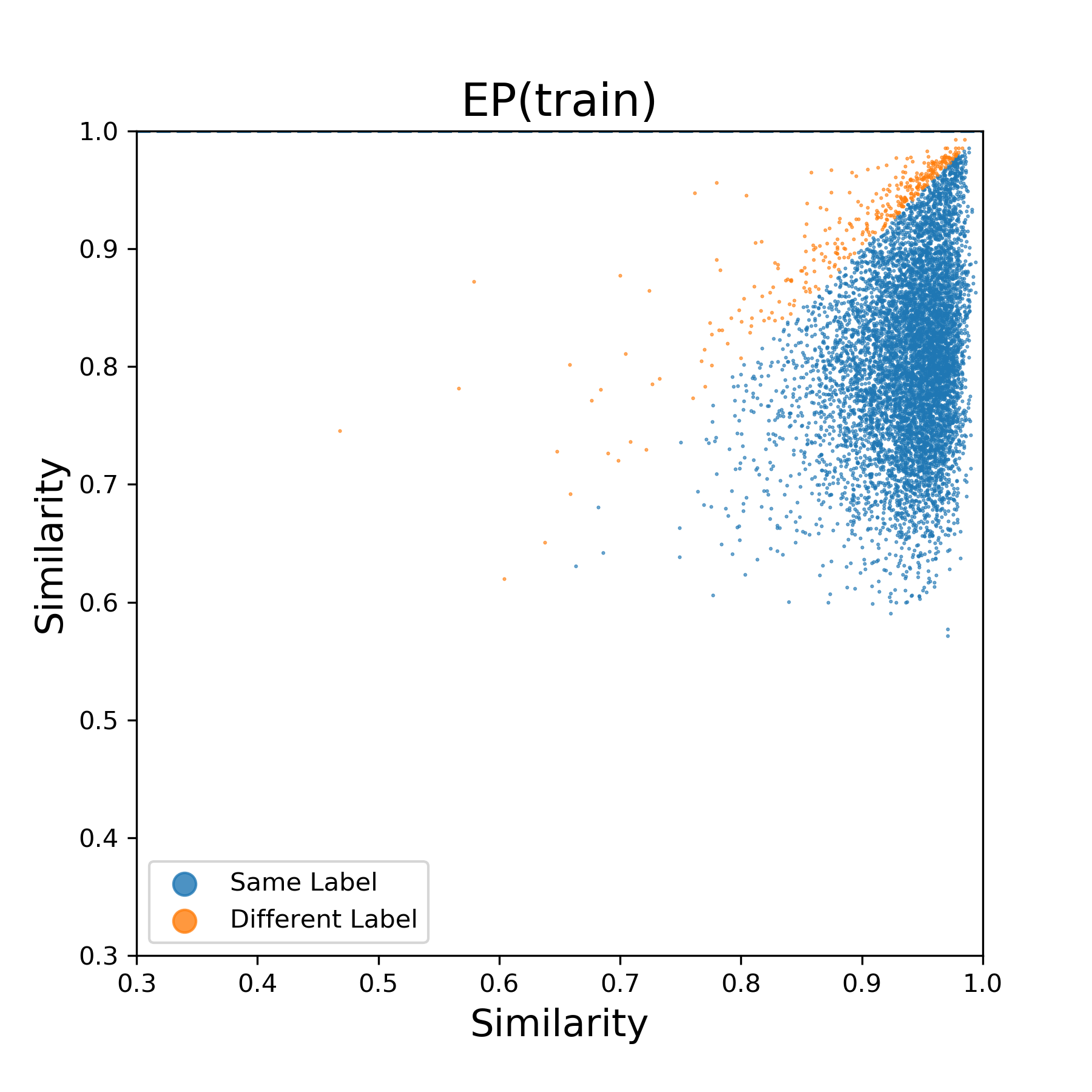}
	\includegraphics[width=0.49\textwidth]{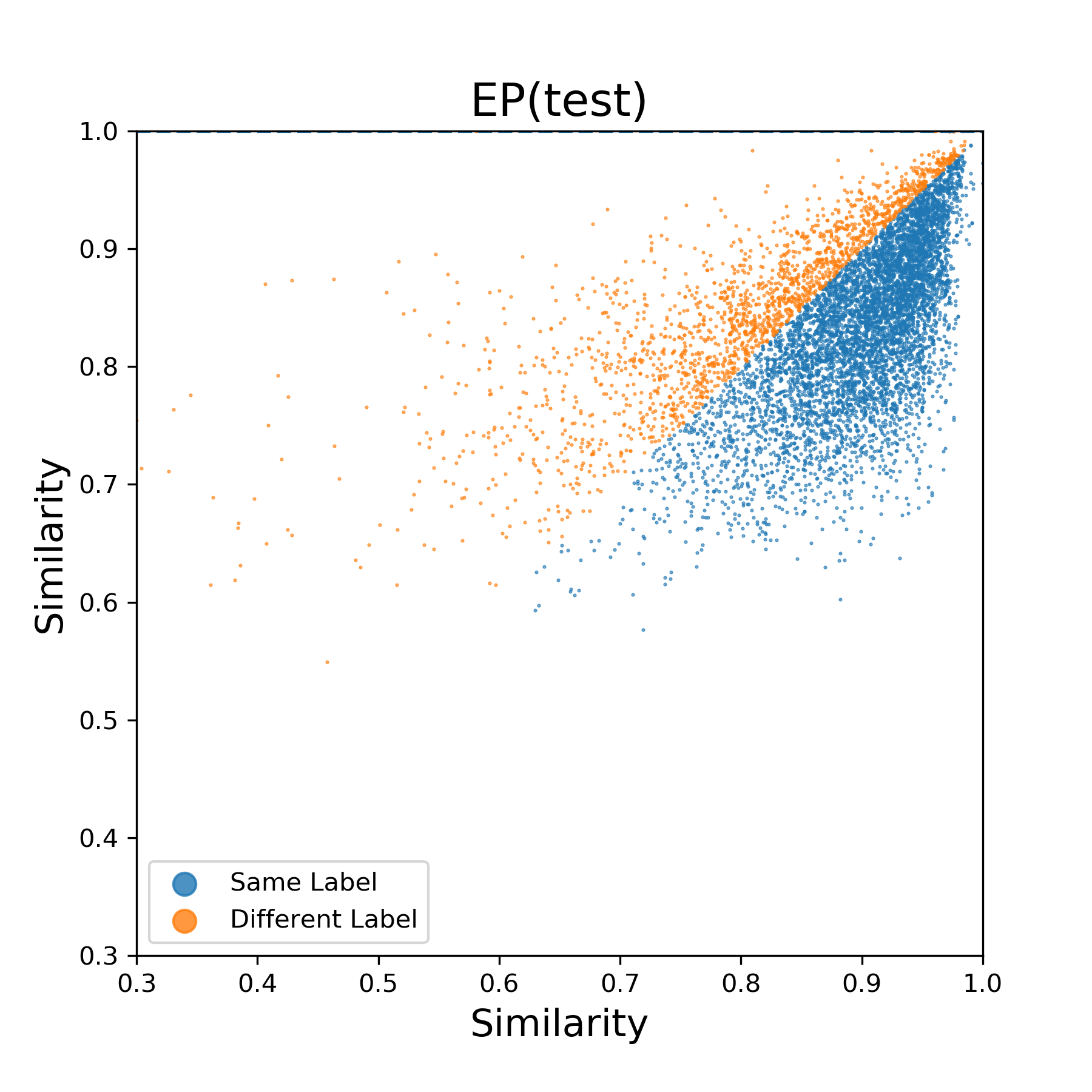}
	\includegraphics[width=0.49\textwidth]{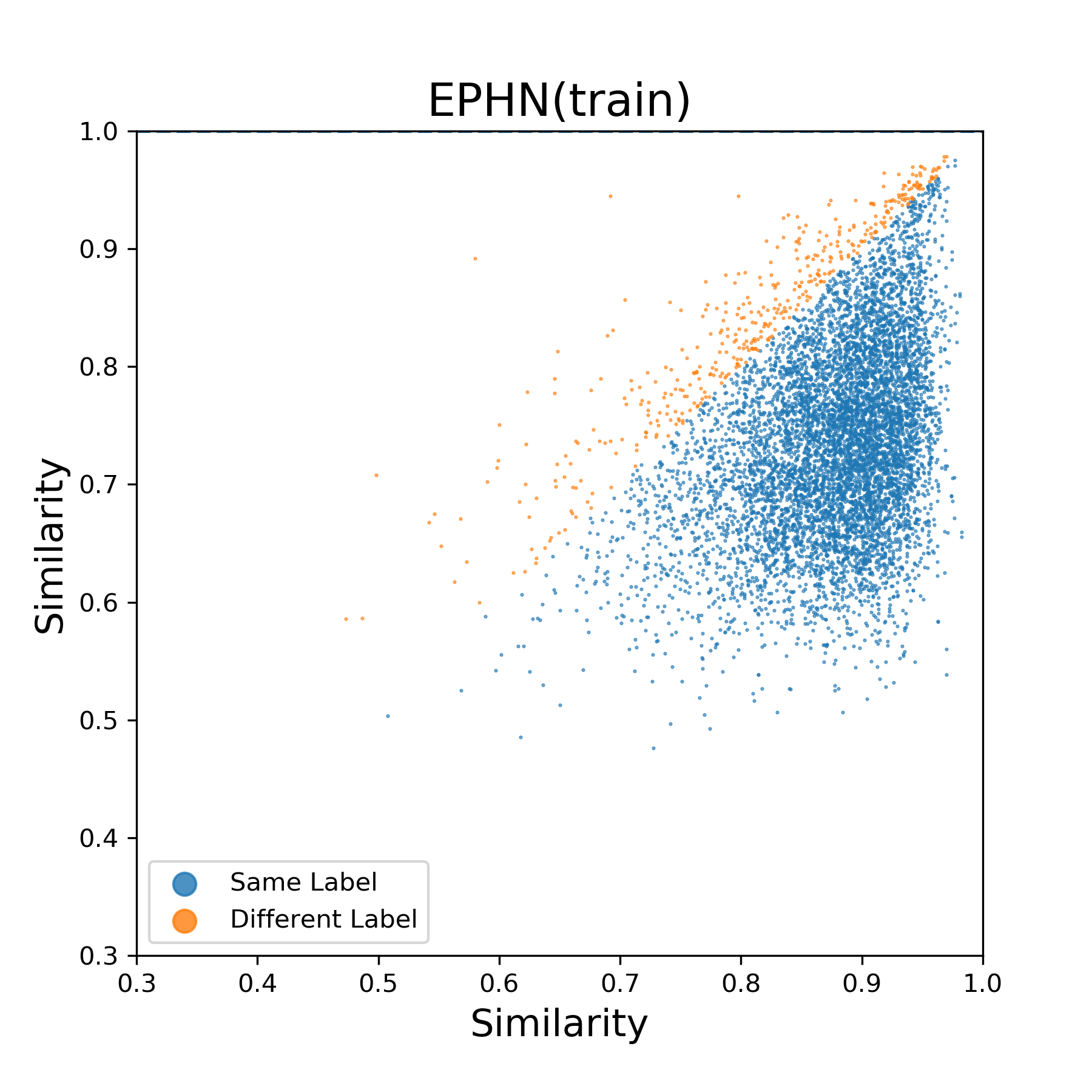}
	\includegraphics[width=0.49\textwidth]{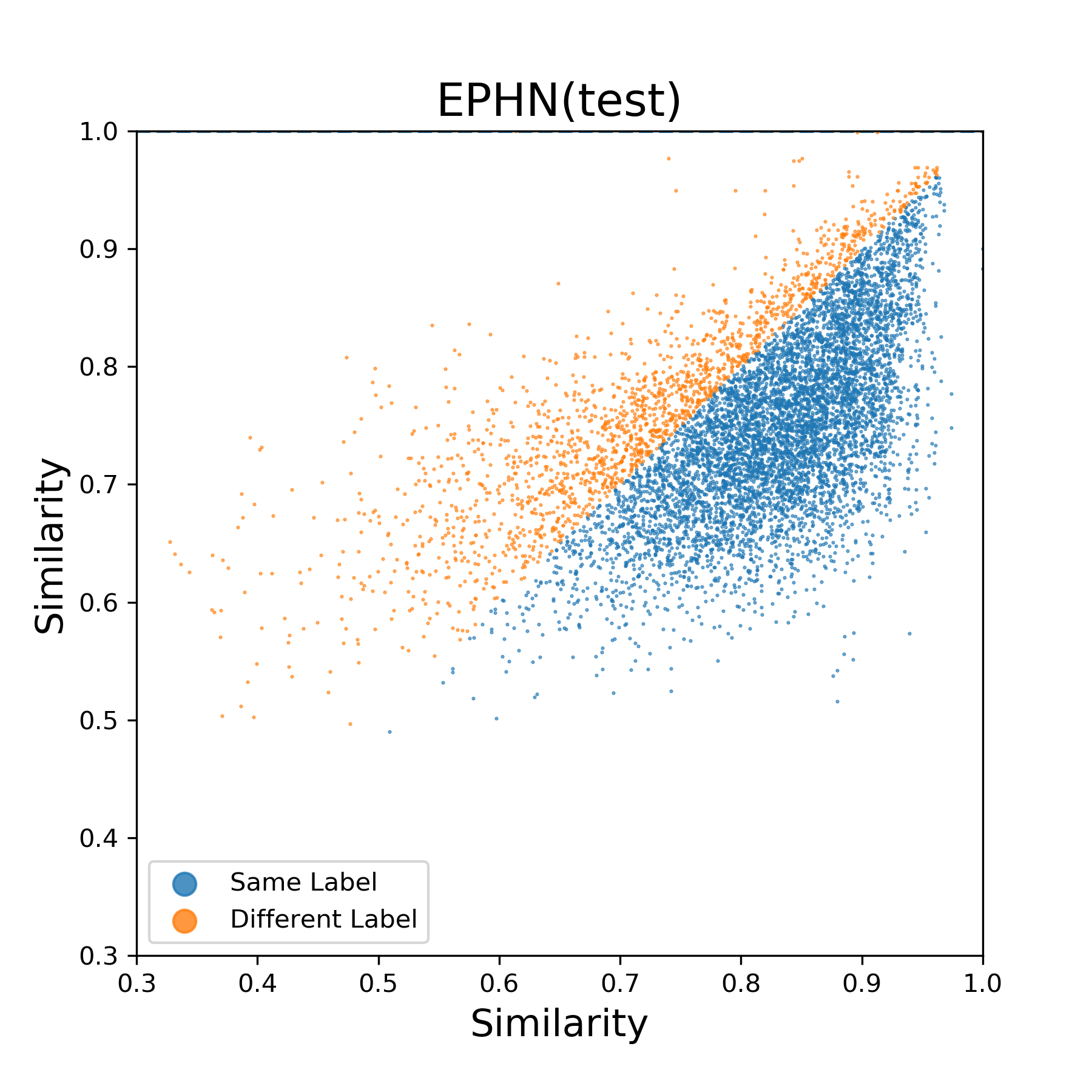}
	\includegraphics[width=0.49\textwidth]{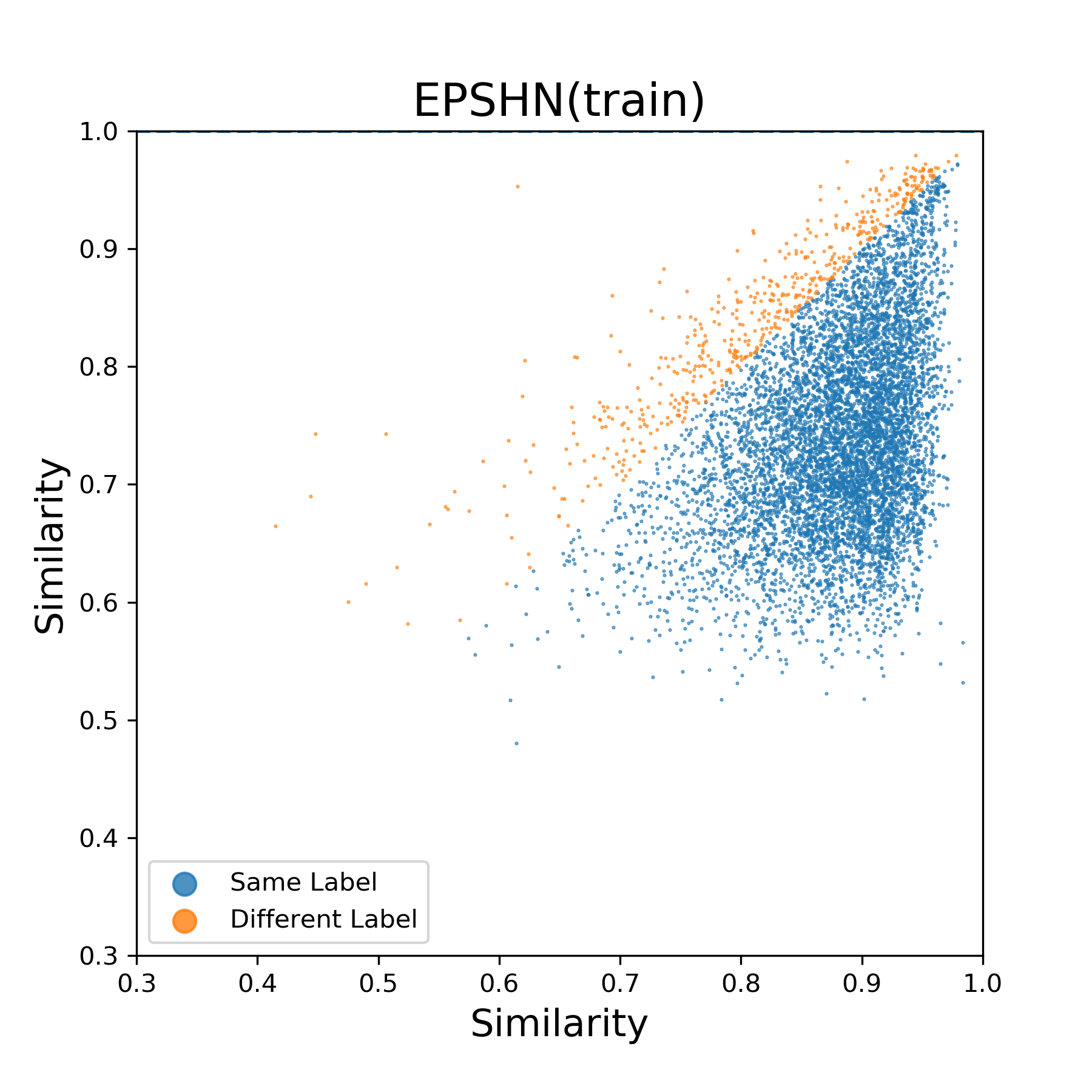}
	\includegraphics[width=0.49\textwidth]{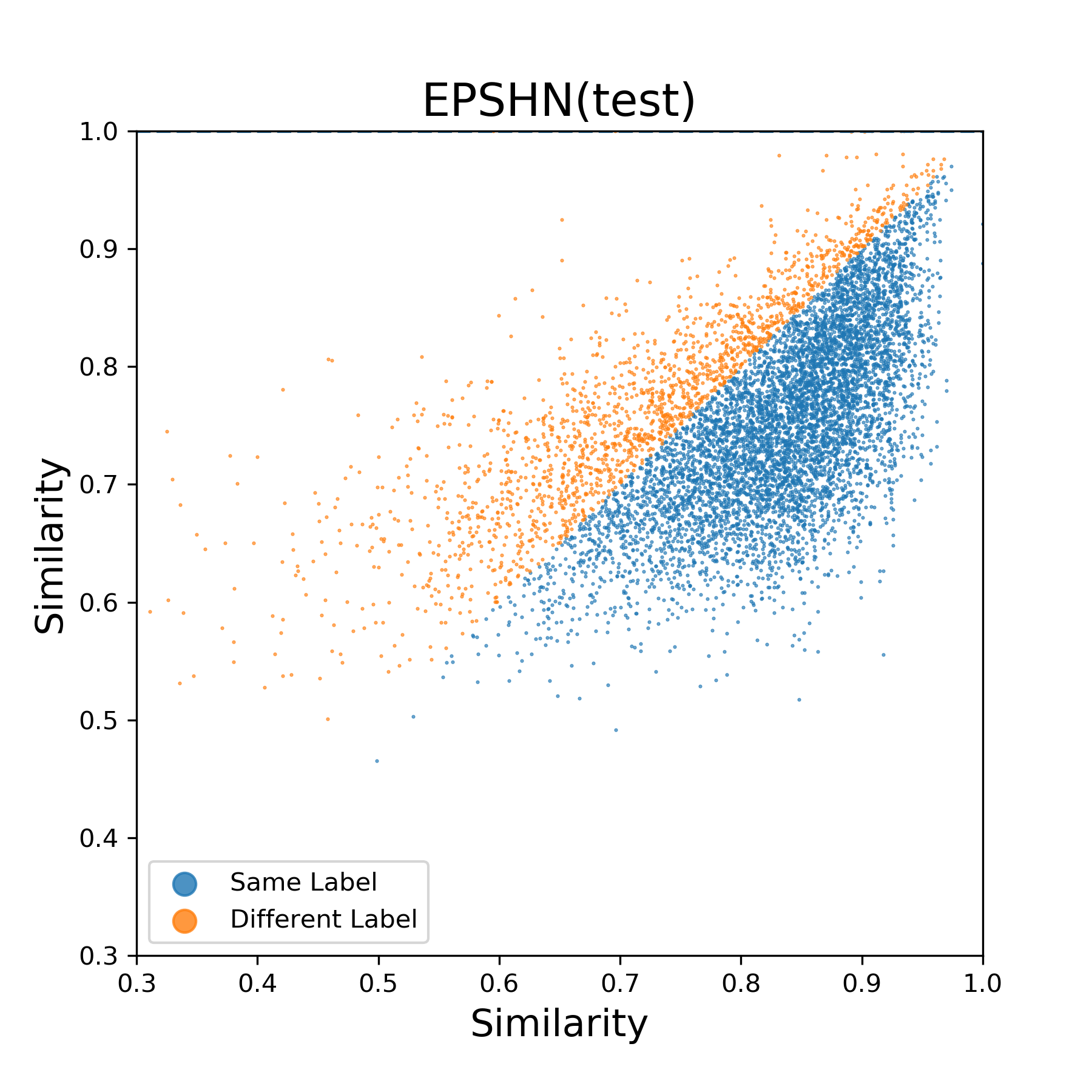}
	\end{subfigure}
\caption{A visual summary of the embedding structure for Batch All (BA), Hard Positive (HP), N-Pairs, and 3 approaches that focus on Easy Positives. Each figure shows the similarity (in embeddings space) of the closest same class image (on the x-axis) and the closest different class image (on the y-axis).  It is especially interesting to note that N-Pairs is very good on training data (similarity to closest same class image is mostly greater than 0.9), but this does not generalize to embedding unseen classes. Embeddings trained with easy-positives are more spread out (the closest positives are not as close), but generalize better to new classes.}
	\label{fig:simdistr}
\end{figure*}

%% file: figs/tsne.tex
\begin{figure*}[t]
    \centering
	\begin{subfigure}[b]{.4\textwidth}
	\centering
	\includegraphics[width=0.8\textwidth]{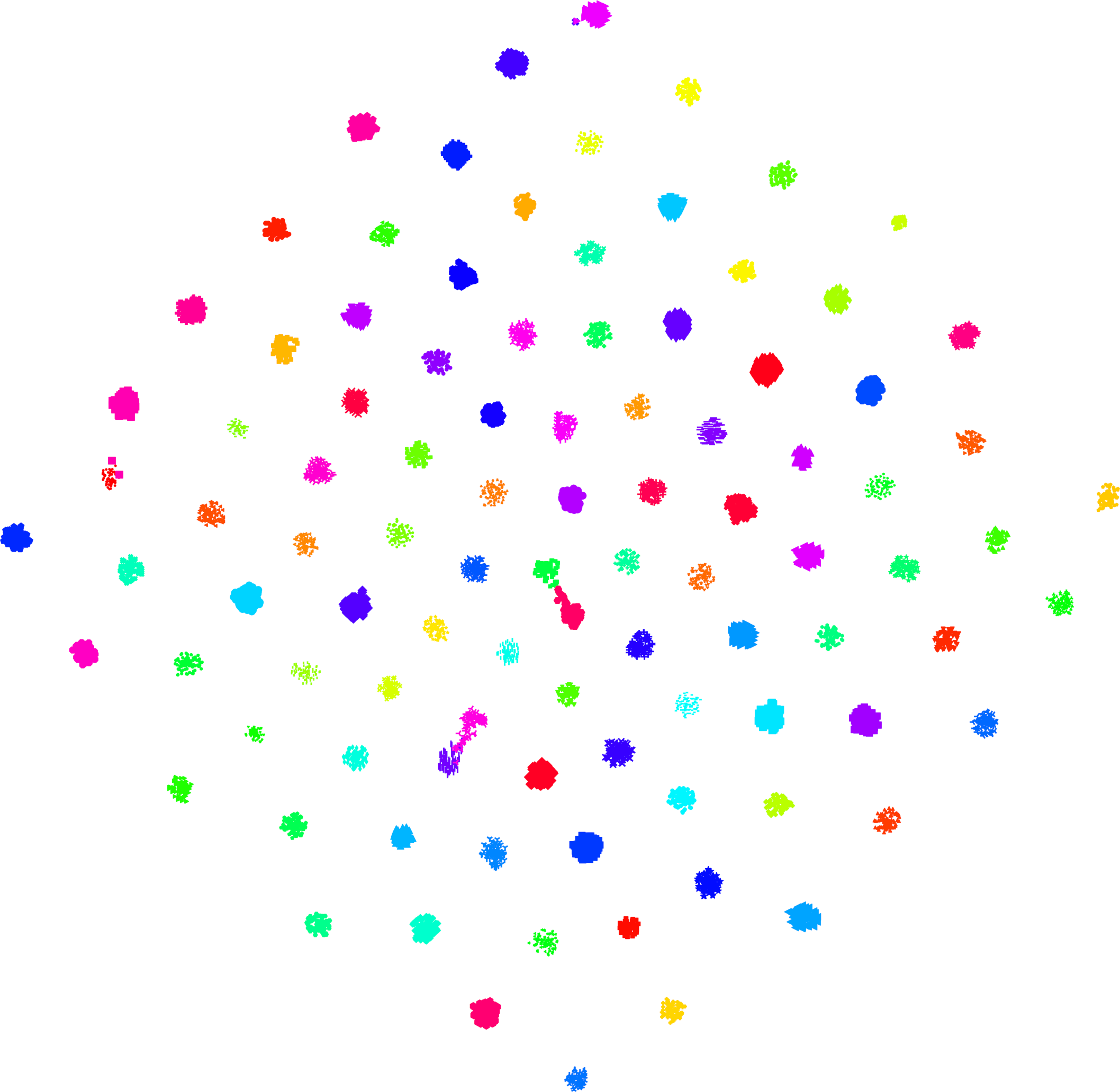}
	\caption{N-pair training embedding}
	\end{subfigure}
    \begin{subfigure}[b]{.4\textwidth}
    \centering
	\includegraphics[width=0.8\textwidth]{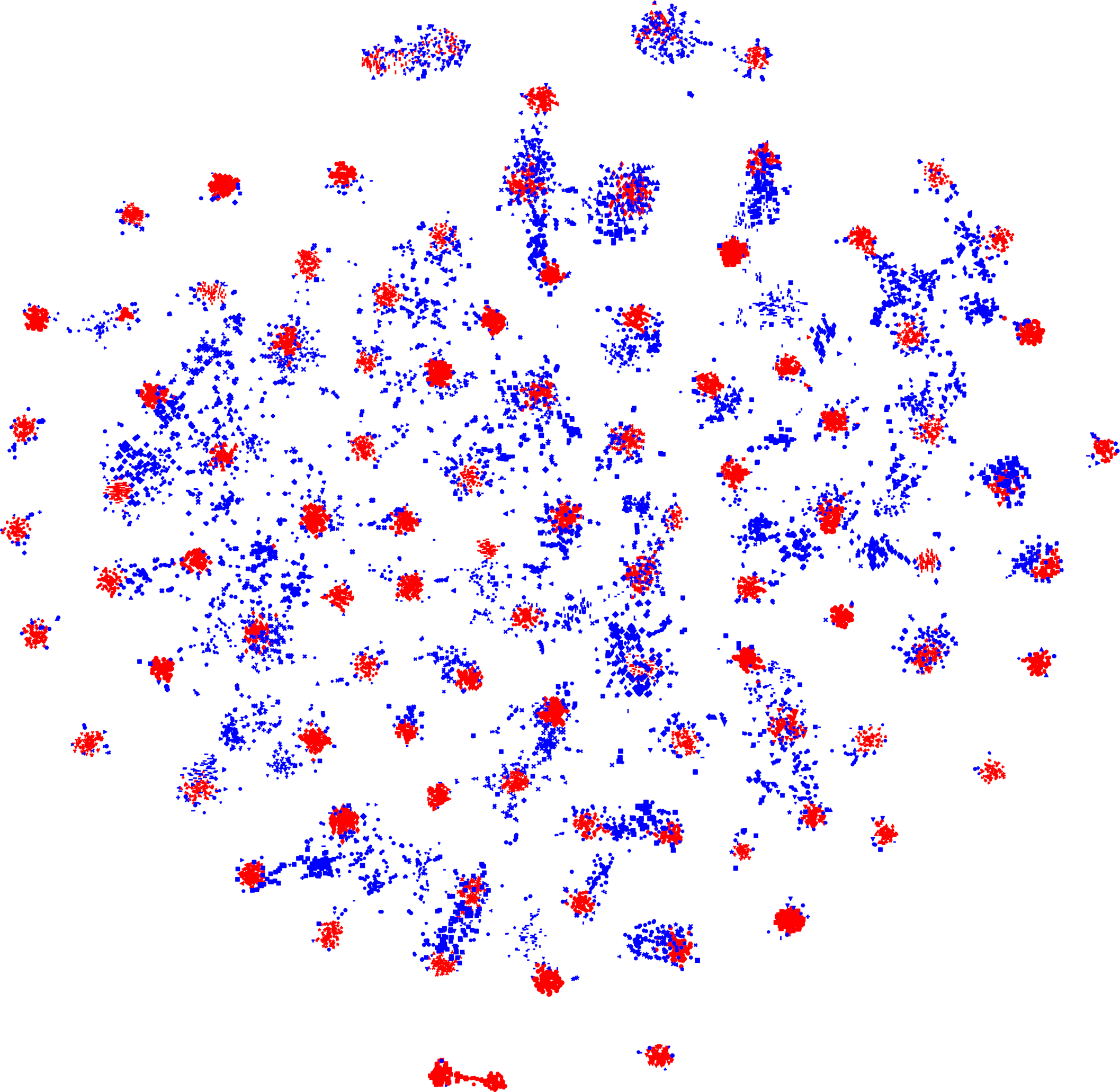}
	\caption{N-pair testing + training embedding}
	\end{subfigure}
	\begin{subfigure}[b]{.4\textwidth}
	\centering
	\includegraphics[width=0.8\textwidth]{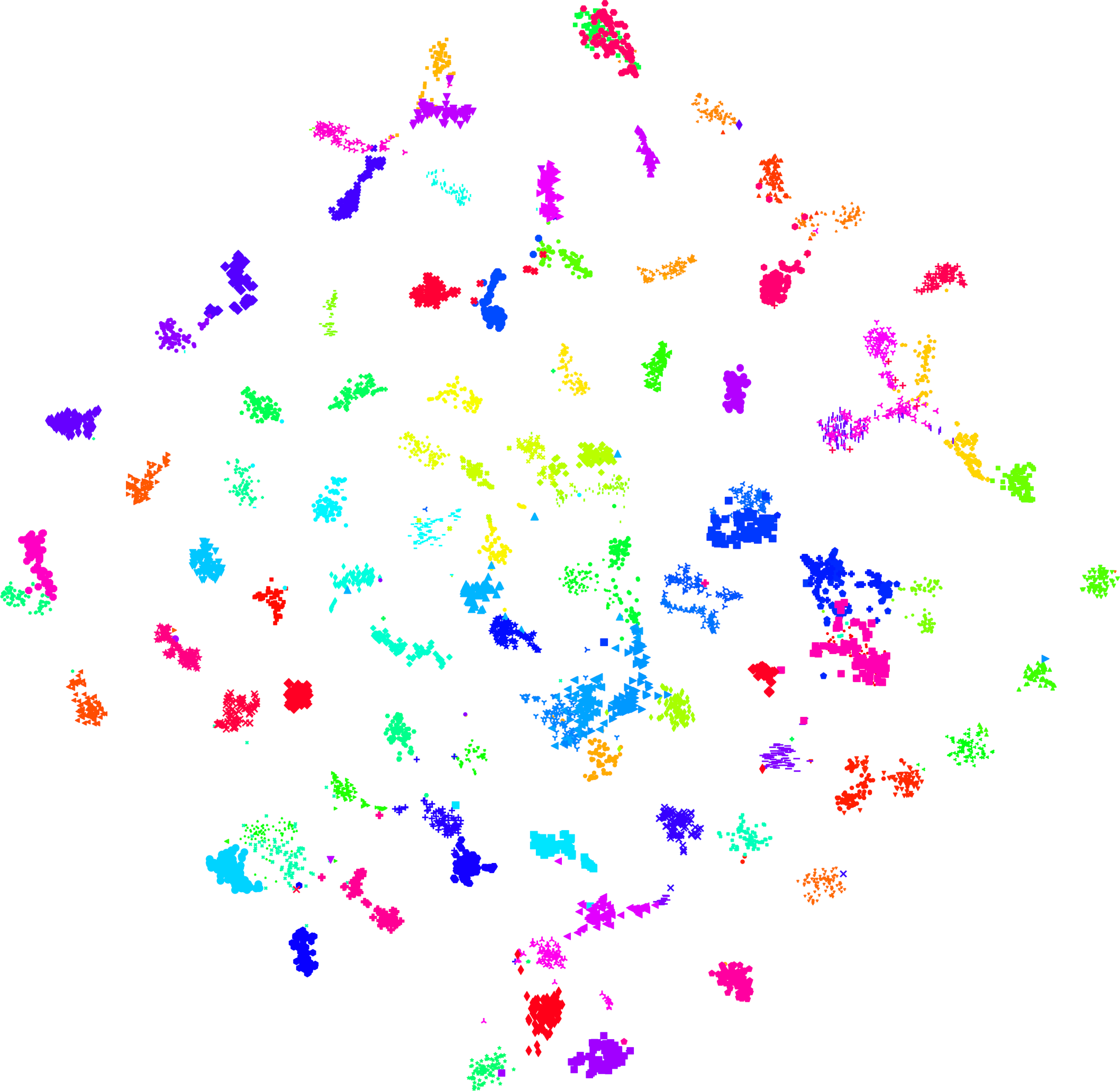}
	\caption{EPSHN training embedding}
	\end{subfigure}
	\begin{subfigure}[b]{.4\textwidth}
	\centering
    \includegraphics[width=0.8\textwidth]{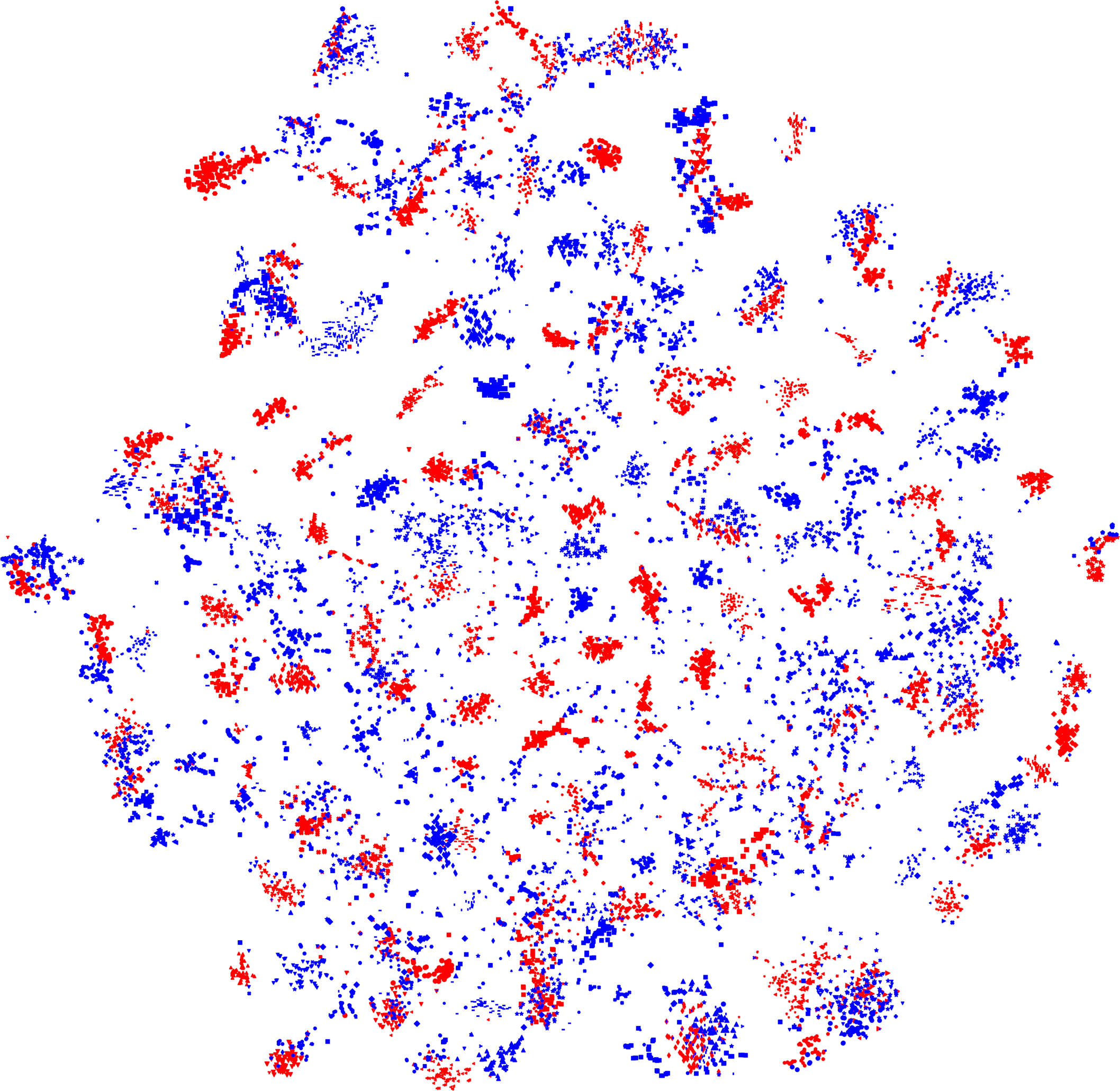}
    \caption{EPSHN testing + training embedding}
    \end{subfigure}
	\caption{A t-SNE visualization of N-pair loss and EPSHN, showing the embedding of the training categories to highlight their structure (left) and a joint embedding of the training and the testing classes to highlight where test data is embedded compared to training data (right).  The EPSHN approach only requires that nearby points are similar, so the classes are not embedding as tightly as clusters compared to the N-Pair approach.  Additionally, with EPSHN, new data is better spread out and mapped less directly on top of a training category, compared the N-pair embedding.}
	\label{fig:tsne}
\end{figure*}

%% file: table/CUBCAR_model.tex
\begin{table*}[t]
\begin{center}
\begin{tabular}{|c|cccc|cccc|cccc|}
\hline
Model & \multicolumn{4}{|c|}{GoogleNet} & \multicolumn{4}{|c|}{Resnet18} & \multicolumn{4}{|c|}{Resnet50}\\
\hline
\multicolumn{13}{|c|}{CUB}\\
\hline
Method & R@1 & R@2 & R@4 & R@8 & R@1 & R@2 & R@4 & R@8 & R@1 & R@2 & R@4 & R@8\\
\hline
TRIPLET$^{64}$    
& 42.6 & 55.0 & 66.4 & 77.2 & - & - & - & - & - & - & - & - \\
N-PAIR$^{64}$
& 45.4 & 58.4 & 69.5 & 79.5 & 52.4 & 65.7 & 76.8 & 84.6 & 53.2 & 65.3 & 76.0 & 84.8\\
PROXY$^{64}$
& 49.2 & 61.9 & 67.9 & 72.4 & 51.5 & 63.8 & 74.6 & 84.0 & 55.5 & 67.7 & 78.2 & 86.2\\
EPSHN$^{64}$
& \bf{51.7} & \bf{64.1} & \bf{75.3} & \bf{83.9} & \bf{54.2} & \bf{66.6} & \bf{77.4} & \bf{86.0} & \bf{57.3} & \bf{68.9} & \bf{79.3} & \bf{87.2}\\
\hline

\multicolumn{13}{|c|}{CAR}\\
\hline
Method & R@1 & R@2 & R@4 & R@8 & R@1 & R@2 & R@4 & R@8 & R@1 & R@2 & R@4 & R@8\\
\hline
TRIPLET$^{64}$
& 51.5 & 63.8 & 73.5 & 81.4 
& - & - & - & - 
& - & - & - & - \\
N-PAIR$^{64}$
& 53.9 & 66.8 & 77.8 & 86.4 
& 55.7 & 67.4 & 77.0 & 84.5
& 58.3 & 69.5 & 78.3 & 86.4\\
PROXY$^{64}$
& \bf{73.2} & \bf{82.4} & \bf{86.4} & 88.7 
& 62.2 & 73.0 & 81.6 & 87.9
& 66.2 & 76.9 & 84.9 & 90.5\\
EPSHN$^{64}$
& 66.4 & 76.8 & 85.2 & \bf{90.7} 
& \bf{73.2} & \bf{82.5} & \bf{88.6} & \bf{93.0} 
& \bf{75.5} & \bf{84.2} & \bf{90.3} & \bf{94.2}\\
\hline


\end{tabular}
\end{center}
\caption{Retrieval Performance on the CUB and CAR dataset.}
\label{table:CUBCAR}
\end{table*}

%% file: table/SOTA.tex
\begin{table*}[t]
\setlength{\tabcolsep}{0.4em}
\begin{center}
\begin{tabular}{|c|ccc|ccc|ccc|ccc|}
\hline
Dataset & 
\multicolumn{3}{|c|}{CUB} & \multicolumn{3}{|c|}{CAR} & \multicolumn{3}{|c|}{SOP} & \multicolumn{3}{|c|}{In-shop}\\
\hline
Method & R@1 & R@2 & R@4 & R@1 & R@2 & R@4 & R@1 & R@10 & R@100& R@1 & R@10 & R@20\\
\hline
HDC$^{384}$ 
& 53.6 & 65.7 & 77.0 
& 73.2 & 82.4 & 86.4 
& 69.5 & 84.4 & 92.8 
& 62.1 & 84.9 & 89.0 \\
BIER$^{512}$
& 55.3 & 67.2 & 76.9 
& 78.0 & 85.8 & 91.1 
& 72.7 & 86.5 & 94.0 
& 76.9 & 92.8 & 95.2 \\
HTL$^{512}$
& 57.1 & 68.8 & 78.7 
& 81.4 & 88.0 & 92.7
& 74.8 & 88.3 & 94.8 
& - & - & -\\
ABE$^{512}$
& 60.6 & 71.5 & 79.8 
& 85.2 & 90.5 & 94.0 
& 76.3 & 88.4 & 94.8 
& 87.3 & \bf{96.7} & \bf{97.9} \\
DREML$^{576}$
& 63.9 & 75.0 & 83.1 
& \bf{86.0} &\bf{91.7} & \bf{95.0}
& - & - & -
& - & - & -\\
EPSHN$^{512}$ 
& \bf{64.9} & \bf{75.3} & \bf{83.5} 
& 82.7 & 89.3 & 93.0
& {\bf78.3} & {\bf90.7} & {\bf96.3} 
& \bf{87.8} & 95.7 & 96.8\\
\hline
\end{tabular}
\end{center}
\caption{Retrieval Performance on the CUB, CAR, SOP and In-shop datasets comparing to the best reported results for more complex approaches and/or ensembles.}
\label{table:SOTA}
\end{table*}



%% file: table/Hotels.tex
\begin{table}[t]
\begin{center}
\begin{tabular}{|c|ccc|}
\hline
\multicolumn{4}{|c|}{Hotels-50K}\\

\hline
Method & R@1 & R@10 & R@100 \\
\hline
BATCH-ALL$^{256}$ & 8.1 & 17.6 & 34.8 \\
EPSHN$^{256}$ & \bf{16.3} & \bf{30.5} & \bf{49.9}\\
\hline
\end{tabular}
\end{center}
\caption{Retrieval performance on the Hotels-50K dataset~\cite{hotels50k}, comparing to the author's original results trained with Resnet-50 and Batch All triplet loss.}
\label{table:Hotels}
\end{table}